\documentclass{article}

\usepackage[final]{neurips_2026}
\makeatletter
\renewcommand{\@notice}{}
\makeatother
\usepackage[most]{tcolorbox}
\usepackage{comment}
\usepackage[utf8]{inputenc}
\usepackage[T1]{fontenc}
\usepackage{hyperref}
\usepackage{url}
\usepackage{booktabs}
\usepackage{amsfonts}
\usepackage{nicefrac}
\usepackage{microtype}
\usepackage{xcolor}
\usepackage{graphicx}
\usepackage{float}
\usepackage{tikz}
\usetikzlibrary{calc,positioning}
\usepackage{siunitx}

\definecolor{ink}{HTML}{16191B}
\definecolor{labelgrey}{HTML}{6D7679}
\definecolor{agentink}{HTML}{4C6B7A}
\definecolor{agentedge}{HTML}{C3CDD3}
\definecolor{agentbg}{HTML}{EFF3F5}
\definecolor{patink}{HTML}{7A6A55}
\definecolor{patedge}{HTML}{D6D2CA}
\definecolor{patbg}{HTML}{FFFFFF}
\definecolor{hotline}{HTML}{B4442C}   %
\definecolor{hotfill}{HTML}{FAE9E4}
\definecolor{coolline}{HTML}{2E7D5B}  %
\definecolor{coolfill}{HTML}{E6F2EC}

\title{ Does the Voice Matter? Evaluating Frustration Detection on Deployed
Clinical Voice-Agent Calls}

\author{%
  Annika Michael \\
  Ufonia Limited \\
  London, UK \\
  \texttt{am@ufonia.com}
  \And
  Zachary Ellis \\
  Ufonia Limited \\
  London, UK \\
  \texttt{ze@ufonia.com}
  \And
  Yuxiang Zhou \\
  Ufonia Limited \\
  London, UK \\
  \texttt{yz@ufonia.com}
  \And
  Samuel Thio \\
  King's College London \\
  London, UK \\
  \texttt{samuel.j.thio@kcl.ac.uk}
  \And
  Yajie Vera He \\
  Ufonia Limited \\
  London, UK \\
  \texttt{yh@ufonia.com}
  \And
  Spencer Hazel \\
  Newcastle University \\
  Newcastle upon Tyne, UK \\
  \texttt{spencer.hazel@newcastle.ac.uk}
  \And
  Jared Joselowitz \\
  Ufonia Limited \\
  London, UK \\
  \texttt{jj@ufonia.com}
}
\makeatletter
\usepackage{xcolor}
\definecolor{caserule}{gray}{0.82}
\definecolor{casetint}{gray}{0.975}
\newcommand{\tick}{\textcolor{black!75}{\checkmark}}
\newcommand{\cross}{\textcolor{black!75}{$\times$}}
\newcommand{\casehead}[3]{%
  {\small\textbf{#1}}\hfill{\scriptsize\textsc{#2}\,$\cdot$\,#3\,w}\par
  \vspace{3pt}\hrule height 0.4pt \vspace{5pt}}
\newcommand{\casenote}[1]{{\footnotesize\itshape #1\par}\vspace{3pt}}
\newenvironment{casebox}
  {\par\vspace{4pt}\noindent\begin{minipage}{\linewidth}%
   \setlength{\fboxsep}{7pt}\begin{lrbox}{\@tempboxa}\begin{minipage}{\dimexpr\linewidth-2\fboxsep-2\fboxrule}\footnotesize}
  {\end{minipage}\end{lrbox}%
   \fcolorbox{caserule}{casetint}{\usebox{\@tempboxa}}\end{minipage}\par\vspace{6pt}}
\makeatother

\begin{document}

\maketitle

\begin{abstract}
Autonomous voice agents now conduct routine clinical follow-up calls, and because no
service can review every call by hand, some automated triage is unavoidable. A patient becoming frustrated is one thing that triage should catch. These systems typically read only the transcript the speech-to-text stage produces, not the voice, and whether that is enough is largely untested on real clinical speech. We study per-turn frustration detection on real patient calls from a deployed clinical voice agent: \num{150} turns from six calls labelled by three annotators, with a length-matched acted corpus as a control, and \num{714} turns from 30 calls. Patient turns are overwhelmingly short: across the deployed calls, roughly \SI{59}{\percent} are one or two words, against \SI{13}{\percent} in the full acted corpus, and a corpus accuracy gap remains even after the acted set is length-matched. On one-word turns, transcript-only detection does not exclude chance. Across five audio-capable language models, the transcript flags \num{3} or \num{4} of the \num{39} frustrated one-word turns in the 30 calls, the same three for every model, and a classifier given only the word count reaches \SI{57.3}{\percent}. Whether audio helps depends on the model and on the instruction. Asked to attend to delivery, two Gemini 2.5 models gain \SI{3.0}{pp} over the transcript across 30 calls (\SIrange{-1.0}{7.5}{pp}). Of the turns they confidently clear, \SI{6.9}{\percent} were in fact frustrated for audio against \SI{13.0}{\percent} for the transcript, a difference of \SI{6.1}{pp} (\SIrange{2.2}{10.4}{pp}). Across the 30 calls, given one prompt identical for both inputs, no model's audio--text interval excludes zero. Under the input-specific prompts, a newer Gemini model and two smaller open models do worse with audio than with the transcript (\SI{-3.5}{pp}, \SI{-8.6}{pp} and \SI{-10.6}{pp}). A WavLM probe trained on the clinical calls matches the best prompted model overall (\SI{71.6}{\percent}), but flags none of the \num{18} frustrated one-word turns in the 24 added calls. Absolute rates reflect our sampling, not deployment prevalence. When a patient's turn is short, the transcript alone is not enough to rule out frustration.
\end{abstract}

\section{Introduction}
\label{sec:intro}

Clinical voice agents are now deployed to autonomously conduct routine follow-up calls~\citep{adams2025voiceagents}, making automated triage increasingly necessary. Patient frustration is one state such systems should detect: frustrated patients may disengage, answer tersely, or hang up before clinically important information is said~\citep{klein2002frustration, hernandez-caralt-etal-2025-stupid}. How a patient sounds should change what happens next. A clinician who heard strain in an otherwise fine answer would probe further, and the agent should too. Whether these systems can tell that has happened is largely unknown. We study a per-turn decision: whether each patient utterance is frustrated or not.

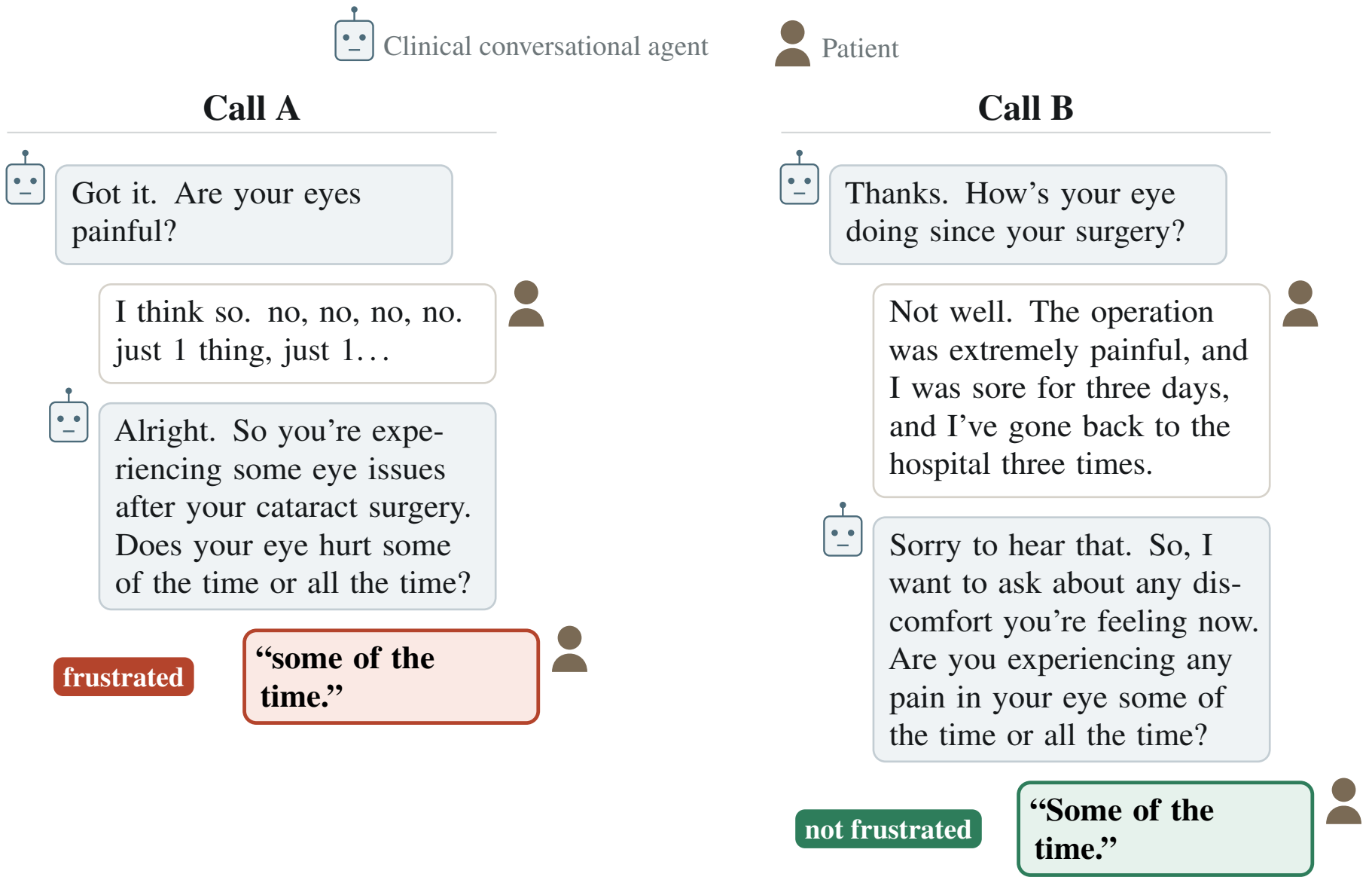
\begin{figure}[htbp]
  \centering
  \begin{tikzpicture}[
  font=\footnotesize,
  node distance=1.6mm,
  pics/patient/.style={code={
    \fill[patink] (0,0.10) circle (0.105);
    \fill[patink] (-0.155,-0.20) .. controls (-0.155,0.015) and (0.155,0.015)
                          .. (0.155,-0.20) -- cycle;
  }},
  pics/agent/.style={code={
    \draw[agentink,line width=0.45pt] (0,0.20) -- (0,0.285);
    \fill[agentink] (0,0.30) circle (0.028);
    \draw[agentink,line width=0.45pt,fill=agentbg,rounded corners=1.4pt]
      (-0.165,-0.145) rectangle (0.165,0.20);
    \fill[agentink] (-0.072,0.055) circle (0.028);
    \fill[agentink] ( 0.072,0.055) circle (0.028);
    \draw[agentink,line width=0.45pt] (-0.05,-0.055) -- (0.05,-0.055);
  }},
  bub/.style={rounded corners=3pt, inner sep=4pt, align=left,
              text width=32mm, line width=0.5pt},
  agentbub/.style={bub, draw=agentedge, fill=agentbg, text=ink, anchor=north west},
  patbub/.style  ={bub, draw=patedge,   fill=patbg,   text=ink, anchor=north east},
  hotbub/.style  ={bub, text width=23mm, draw=hotline,  fill=hotfill,  line width=0.9pt,
                   anchor=north east},
  coolbub/.style ={bub, text width=23mm, draw=coolline, fill=coolfill, line width=0.9pt,
                   anchor=north east},
  pill/.style={rounded corners=2.4pt, inner sep=2.4pt, font=\scriptsize\bfseries,
               text=white, anchor=east},
  head/.style={font=\small\bfseries, text=ink},
  leg/.style={font=\scriptsize, text=labelgrey, anchor=west, inner sep=0pt},
]

\pic at (3.05,0.66) {agent};
\node[leg] at (3.30,0.62) {Clinical conversational agent};
\pic at (6.90,0.66) {patient};
\node[leg] at (7.15,0.62) {Patient};

\node[head] at (2.15,0.10) {Call A};
\draw[labelgrey!30,line width=0.4pt] (0,-0.12) -- (4.3,-0.12);

\node[agentbub] (a1) at (0.42,-0.40)
  {Got it. Are your eyes painful?};
\pic at ($(a1.north west)+(-0.26,-0.20)$) {agent};

\node[patbub] (a2) at ($(a1.south west)+(3.88,-0.16)$)
  {I think so. no, no, no, no. just 1 thing, just 1\ldots};
\pic at ($(a2.north east)+(0.26,-0.18)$) {patient};

\node[agentbub] (a3) at ($(a2.south west)+(-0.0,-0.16)$)
  {Alright. So you're experiencing some eye issues after your cataract surgery. Does your eye hurt some of the time or all the time?};
\pic at ($(a3.north west)+(-0.26,-0.20)$) {agent};

\node[hotbub] (a4) at ($(a3.south west)+(3.88,-0.16)$) {\bfseries ``some of the time.''};
\pic at ($(a4.north east)+(0.26,-0.18)$) {patient};
\node[pill,fill=hotline] at ($(a4.west)+(-0.42,0)$) {frustrated};

\node[head] at (8.95,0.10) {Call B};
\draw[labelgrey!30,line width=0.4pt] (6.8,-0.12) -- (11.1,-0.12);

\node[agentbub] (b1) at (7.22,-0.40)
  {Thanks. How's your eye doing since your surgery?};
\pic at ($(b1.north west)+(-0.26,-0.20)$) {agent};

\node[patbub] (b2) at ($(b1.south west)+(3.88,-0.16)$)
  {Not well. The operation was extremely painful, and I was sore for three days, and I've gone back to the hospital three times.};
\pic at ($(b2.north east)+(0.26,-0.18)$) {patient};

\node[agentbub] (b3) at ($(b2.south west)+(-0.0,-0.16)$)
  {Sorry to hear that. So, I want to ask about any discomfort you're feeling now. Are you experiencing any pain in your eye some of the time or all the time?};
\pic at ($(b3.north west)+(-0.26,-0.20)$) {agent};

\node[coolbub] (b4) at ($(b3.south west)+(3.88,-0.16)$) {\bfseries ``Some of the time.''};
\pic at ($(b4.north east)+(0.26,-0.18)$) {patient};
\node[pill,fill=coolline] at ($(b4.west)+(-0.30,0)$) {not frustrated};

\end{tikzpicture}
  \caption{Two real calls with similar questions and responses. The patient's response ``Some of the time'' is identical in wording but annotators labelled it frustrated in Call A and not frustrated in Call B. Transcript excerpts are verbatim.}
  \label{fig:cand3}
\end{figure}

The obvious way to build that triage is on the transcript: cheap, already produced by the speech-to-text stage, and on standard benchmarks about as good as audio~\citep{pang2026voxparadox, chen-etal-2026-audio, selvakumar2025multivox}. Our own headline initially agrees: pooled across three models, transcript-only and audio-only accuracy differ by about a point. But that average misleads. Real clinical speech is dominated by short turns: roughly \SI{59}{\percent} of patient turns are one or two words, leaving little lexical evidence while preserving the delivery. The clearest case is the same word carrying opposite states: in Figure~\ref{fig:cand3}, an identical ``Some of the time'' is labelled frustrated in one call and not frustrated in another, so a transcript-only system cannot tell them apart and clears both. The two inputs therefore do not fail on the same turns: transcript-only detection silently clears short frustrated turns where a triage system can least afford it. The pooled average also hides a split by model and by instruction: audio helps the more capable models only when they are asked to listen for delivery, and hurts a weaker one either way.

This is structural. The deployed agent is cascaded: speech is transcribed, a language model reasons over the text, and a reply is synthesised. Paralinguistic cues such as hesitation, strain, and pauses are therefore discarded before reasoning~\citep{miller2026counterfactual, liu2026hear2act}. The transcript is what the agent uses today; audio represents what would be available if that delivery were retained.

The same failure is much less visible on the acted emotion corpora the field validates against, because they contain few short turns: \SI{13}{\percent} of IEMOCAP's utterances are one or two words, against \SI{59}{\percent} of patient turns on deployed calls. Existing evaluations rarely test this on real deployed clinical speech. We therefore compare transcript-only and direct-audio frustration detection on real calls from a deployed clinical voice agent, with turn-level labels agreed by three annotators and a length-matched acted corpus as the control. We study frustration because it is a single, actionable state: audible, tied to a patient disengaging, and mapped to a clear response (probe, soften, or escalate).

\paragraph{Contributions}
\begin{itemize}
  \item \textbf{Real deployed clinical speech is a different distribution from acted benchmarks.}
    Patient speech from real clinical voice-agent calls occupies a different distribution from the
    acted corpora used to benchmark emotion recognition: across the full set of deployed calls,
    roughly \SI{59}{\percent} of patient turns
    are one or two words, against \SI{13}{\percent} in the full acted corpus, and a corpus
    accuracy gap remains even after the acted set is matched on length (Section~\ref{sec:speech}).
  \item \textbf{Transcript-only detection collapses on short turns, which dominate clinical speech.} We give the first evidence on deployed clinical speech that transcript-only frustration detection fails to exclude chance on one-word turns: across three models, no one-word transcript interval clears \SI{50}{\percent}, recall on \textsc{frustrated} one-word clinical turns is \SI{9.5}{\percent} for every model (Table~\ref{tab:oneword}), and a classifier using only word count reaches \SI{58.0}{\percent} (Section~\ref{sec:c2}).
  \item \textbf{Whether audio helps depends on the model.}
    Asked to attend to delivery, the two frontier models improve with audio (\SI{4.0}{pp} on the clinical calls,
    \SIrange{-2.3}{10.3}{pp}; \SI{4.9}{pp}, \SIrange{0.9}{9.1}{pp} with the acted control pooled in), while a smaller
    open model is worse with audio (\SI{-12.7}{pp}, \SIrange{-22.0}{-3.3}{pp}; Section~\ref{sec:complementary}).
  \item \textbf{The findings hold with more calls and more models.} Across 30 calls and two added models, every transcript still flags only \num{3} or \num{4} of the \num{39} frustrated one-word turns, and the frontier pair's transcript still wrongly clears more frustrated turns than its audio (\SI{13.0}{\percent} against \SI{6.9}{\percent}). The frontier audio gain stays positive but its interval includes zero, and Gemini 3.8 Flash and both small open models are worse with audio (Section~\ref{sec:thirty}).
  \item \textbf{The instruction decides whether audio helps.} Given one prompt identical for both inputs, the frontier gain disappears (\SI{-1.3}{pp} and \SI{0.7}{pp} on the six calls), and across the 30 calls no model's audio--text interval excludes zero. Gemini 2.5 Pro's audio flags \num{4} of the \num{39} frustrated one-word turns in place of \num{23}, so audio recovers short frustrated turns only when the model is asked to listen for delivery (Section~\ref{sec:instruction}).
  \item \textbf{A trained acoustic probe is not a fix for short turns.} WavLM probes trained on the 30 calls match the best prompted model overall, but their one-word recall on the six calls does not carry to the other 24 (Section~\ref{sec:probes}).
\end{itemize}

\section{Related work}
\label{sec:related}

\paragraph{Speech emotion recognition, on acted and synthetic data.}
Identifying a speaker’s emotional state from how something is said, rather than what is said, is long established, but its evidence base is dominated by acted, read, and synthetic speech. IEMOCAP \citep{busso2008iemocap} remains the reference set, and RAVDESS \citep{livingstone2018ravdess}, CREMA-D \citep{cao2014cremad} and the MSP corpora \citep{busso2017mspimprov} share its design of prompted, categorical, high-arousal emotion recorded in studio. Self-supervised speech encoders such as WavLM \citep{chen2022wavlm}, which our trained probes build on, are themselves benchmarked for emotion recognition on IEMOCAP. EmoNet-Voice \citep{schuhmann2025emonetvoice} grows the label set rather than the source, with forty categories over a fully synthetic voice set, and reports that agreement with human raters falls away outside high-arousal states. Recent large-scale audio evaluations likewise rely on generated speech: LALM-as-a-Judge \citep{ivry2026lalmjudge} resynthesises dialogues, while AudioJudge \citep{manakul2025audiojudge} and SpeechJudge \citep{zhang2025speechjudge} use read or synthetic speech. Recent work using real audio \citep{sayyad2026reliability} instead evaluates audio quality rather than affect.

\paragraph{Audio LLMs perceive delivery but often privilege the words.}
On controlled benchmarks, audio LLMs fall back on the transcript when words and delivery disagree: VoxParadox \citep{pang2026voxparadox} and LISTEN \citep{chen-etal-2026-audio} report failures when affect is carried mainly by delivery, with related failures in deployed systems missing distress and sarcasm \citep{bartelds2026realtime} and in an audio judge that collapses on single items \citep{miller2026counterfactual}. Hear2Act \citep{liu2026hear2act} shows the signal is present but underused: forcing the model to describe the state in text before acting substantially improves downstream behaviour. Reliance on audio also varies across models \citep{selvakumar2025multivox}. These studies use synthetic, acted or adversarial speech rather than real clinical calls, and do not examine utterance length as an explanatory variable. Whether a newer or larger model in the same family listens better is rarely tested; we compare two generations of one family and two small open models on the same turns.

\paragraph{Utterance length as a confound.}
Prior work has treated length or duration as a predictive feature, identified sentence length as an annotation artefact, and shown that very short patient utterances are particularly difficult for ASR \citep{schmitt2009nomatch, gururangan2018annotation, zolnoori2024asr}. Yet length is rarely controlled when comparing text- and audio-based affect detection. We instead measure the clinical length distribution directly, include a length-only baseline, and construct a length-matched acted control. Self-supervised speech encoders such as WavLM give a trained alternative to prompting, and we test whether a probe on them trained on clinical speech generalises across calls.

\paragraph{Frustration on real clinical speech.}
Clinical voice-agent evaluation has largely focused on what was said rather than how it was delivered, even as end-to-end audio models are known to degrade over multi-turn dialogue \citep{tam2026contexttrap}. Paralinguistic work in healthcare has examined synthetic voice bias \citep{tam2025medvoicebias}, while models targeting frustration are typically evaluated on acted corpora \citep{kim-etal-2026-aligning}. The closest clinical neighbours each miss one axis: CBT-Audio \citep{hu2026cbtaudio} compares audio, text and fusion for distress, but the patient speech is acted; thymia \citep{thymia2026} uses real clinical audio, but read speech and audio-only prediction; and WER-is-unaware \citep{ellis2026wer} studies real deployed cataract-call audio but only through its transcript, identifying as an open problem a patient whose words indicate agreement while the voice conveys something else. To our knowledge, no prior work tests whether audio improves frustration detection over transcripts on spontaneous speech from deployed clinical calls.

\section{Data and annotation}
\label{sec:dataset}

\paragraph{Source.} 

We use two data sources: real patient turns from clinical calls and acted dialogue. Patient turns are drawn from recorded follow-up calls conducted by a deployed clinical voice agent on an ophthalmology pathway. All audio is telephone-band (\SI{8}{\kilo\hertz}) spontaneous speech. Calls potentially containing frustration were selected from over 2000 calls using low patient satisfaction ratings and explicit mentions of not wanting to speak to an automated clinician. At the start of every call, patients are asked whether their data may be used for research, and only calls from patients who consented are included. The recordings are anonymised production calls from a proprietary internal dataset that cannot be shared, and contain no personally identifying information. We obtained IEMOCAP \citep{busso2008iemocap} through the official University of Southern California distribution under its research-use licence, using only the audio, transcripts, and emotion annotations. IEMOCAP serves as the acted benchmark and consists of scripted and improvised interactions elicited from actors in a controlled setting. Its emotion categories were binarised by mapping anger and frustration to \textsc{frustrated}, and happiness, sadness, neutrality, excitement, fear, surprise, disgust, and other to \textsc{not\_frustrated}.

\paragraph{Annotation.} Each patient turn is labelled \textsc{frustrated},
\textsc{not\_frustrated} or \textsc{inconclusive}, the last reserved for poor audio quality or genuine ambiguity. Annotators could listen to surrounding turns for context. Three annotators labelled each turn independently, with final labels determined by majority vote. The dataset comprises 150 patient turns from real calls (75 \textsc{frustrated}, 75 \textsc{not\_frustrated}) and 150 actor turns from IEMOCAP (76 \textsc{frustrated}, 74 \textsc{not\_frustrated}).

\begin{table}[H]
  \caption{Inter-annotator agreement on the clinical and acted evaluation sets. Pairwise raw agreement and Cohen's $\kappa$ are reported alongside overall unanimity and unanimity by utterance length. Inconclusive judgements are excluded from the relevant calculation rather than counted as disagreement.}
  \label{tab:agreement}
  \centering
  \small
  \begin{tabular}{@{}lrrcrr@{}}
    \toprule
    & \multicolumn{2}{c}{Clinical} & & \multicolumn{2}{c}{Acted} \\
    \cmidrule(lr){2-3}\cmidrule(lr){5-6}
    & Raw & $\kappa$ & & Raw & $\kappa$ \\
    \midrule
    Annotators 1--2 & 89.1\% & 0.78 & & 88.3\% & 0.76 \\
    Annotators 1--3 & 83.4\% & 0.67 & & 85.9\% & 0.72 \\
    Annotators 2--3 & 86.2\% & 0.73 & & 89.2\% & 0.78 \\
    \midrule
    \textbf{Unanimous} & \textbf{78.5\%} & & & \textbf{82.0\%} & \\
    \midrule
    \multicolumn{6}{@{}l}{\textit{Unanimous, by utterance length}} \\
    \midrule
    1 word  & 75.0\% & & & 84.8\% & \\
    2--3    & 77.3\% & & & 84.6\% & \\
    4--10   & 82.5\% & & & 88.9\% & \\
    11+     & 78.8\% & & & 73.7\% & \\
    \bottomrule
  \end{tabular}
\end{table}

\paragraph{Agreement is comparable across the two corpora.} All \num{150} acted turns carry three independent judgements. Agreement and unanimity are close between the corpora (Table~\ref{tab:agreement}).

\paragraph{Short turns are not harder for annotators.} On the clinical calls, unanimity is \SI{75.0}{\percent} on one-word turns and \SI{78.8}{\percent} on turns of eleven words or more (Table~\ref{tab:agreement}), giving little indication that short turns are intrinsically harder to label. Human listeners resolve a single-word turn about as reliably as a long one, so the short-turn failures we report later are not an artefact of noisier labels there. Overall annotator unanimity was \SI{78.5}{\percent}, providing a human-agreement reference for model performance.

\paragraph{Evaluation set.} Sections~\ref{sec:c1}--\ref{sec:c3} use \num{150} turns from six calls labelled by all three annotators: \num{75} frustrated and \num{75} not
frustrated, so that guessing confers no advantage and chance is \SI{50}{\percent} by construction. A
matched \num{150}-turn set (76 frustrated, 74 not frustrated) from an acted corpus, rebuilt to the clinical length profile and re-validated by the same three annotators, serves as the control in
Section~\ref{sec:controls}. The acted dataset used utterances from 45 of the 151 available IEMOCAP dialogues.

\paragraph{Extension to 30 calls.} Sections~\ref{sec:thirty}--\ref{sec:probes} use a larger clinical set. We added \num{24} calls from the same deployed agent and pathway. Annotator~1 labelled every patient turn in them on the same criteria as the original six calls. Turns labelled inconclusive are excluded, leaving \num{564} turns, of which \num{65} (\SI{11.5}{\percent}) are frustrated. No turns were dropped to balance the classes. Together with the \num{150} balanced turns from the six calls, the full clinical set has \num{714} turns from 30 calls, \num{140} of them frustrated (\SI{19.6}{\percent}).

\section{Experimental setup}
\label{sec:setup}

\textbf{Models.} We evaluate five multimodal models that accept audio directly, with no fine-tuning or task-specific training. Gemini 2.5 Flash, Gemini 2.5 Pro and Gemini 3.8 Flash are accessed via Vertex AI. The two open models are Qwen2.5-Omni 7B~\citep{xu2025qwen25omni}, served locally through llama.cpp at Q8.0 quantisation, and Voxtral Mini 3B~\citep{liu2025voxtral}. All models use temperature 1.0, the Gemini 2.5 default. We keep it for every model rather than decode greedily, because greedy decoding makes repeats identical and so hides run-to-run variation instead of removing it. Repeats then measure that variation (Appendix~\ref{app:config}). Other sampling parameters remain at provider defaults. Appendix~\ref{app:serving} gives how each model is accessed and called. The acted corpus is scored with Gemini 2.5 Flash, Gemini 2.5 Pro and Qwen2.5-Omni only, so Sections~\ref{sec:c1}--\ref{sec:c3} report those three.

\textbf{Conditions.} Each turn was presented twice, as audio and as transcript only, under two prompt conditions (Appendix~\ref{app:prompts}). The input-specific prompts, used for the main results, direct the audio condition to delivery and the transcript condition to the words. The identical prompt gives both inputs the same instruction, word for word. Every model was run under both conditions on all 30 calls. The model saw the patient turn in isolation, unlike the annotators (Section~\ref{sec:dataset}). Two further experiments on the six calls add context: up to six preceding patient turns, and, separately, the agent's preceding turn, which the patient's turn answers (Section~\ref{sec:whytext}). Responses comprised a label constrained to \textit{frustrated} or \textit{not frustrated} and a free-text rationale used for qualitative analysis only.

\textbf{Trained probes.} As a trained baseline, we fit probes on frozen WavLM Base+~\citep{chen2022wavlm}: one intermediate layer (of layers 6--9 of 12), pooled over the turn, with a logistic regression or an RBF support vector machine on top. One set is trained only on acted IEMOCAP speech (Appendix~\ref{app:probes}), the other on the 30 clinical calls, split into five folds by patient and scored out of fold (Appendix~\ref{app:probethirty}). Layer and pooling are chosen by cross-validation inside each training fold, and three split seeds give three predictions per turn, combined by majority as for the language models.

\textbf{Metrics.} Every turn is classified three times independently in each condition, and the turns' prediction is the majority of those three. Balanced accuracy, the mean of recall and specificity, is reported throughout. Turn length is measured in words after stripping bracketed annotation markers, and bucketed as 1, 2–3, 4–5, 6–10, 11–20 and 21 or more words. The six-call clinical set and the acted control are balanced overall, but the turn length buckets we analyse within them are not, and plain accuracy inside an imbalanced bucket rewards predicting whichever class dominates it. Chance is \SI{50}{\percent} by construction. A single balanced accuracy gets a \SI{95}{\percent} Wilson interval and a difference between conditions a \num{10000}-draw bootstrap over turns, both on the majority-vote predictions. Redrawing each turn's three repeats inside the bootstrap as well moves no bound by more than \SI{2}{pp} and changes no conclusion except the acted frontier pair, whose interval then just includes zero (Appendix~\ref{app:config}). These intervals are used in Sections~\ref{sec:c1}--\ref{sec:c3} and their appendices. Sections~\ref{sec:thirty}--\ref{sec:probes} and Appendix~\ref{app:thirty} instead use a \num{10000}-draw bootstrap over whole calls, because turns from one call are not independent.

\section{Results}
\label{sec:results}
\subsection{Real clinical speech is a different distribution from acted benchmarks}
\label{sec:c1}
\label{sec:speech}
\label{sec:controls}

\begin{figure}[H]
  \centering
  \includegraphics[width=0.7\linewidth]{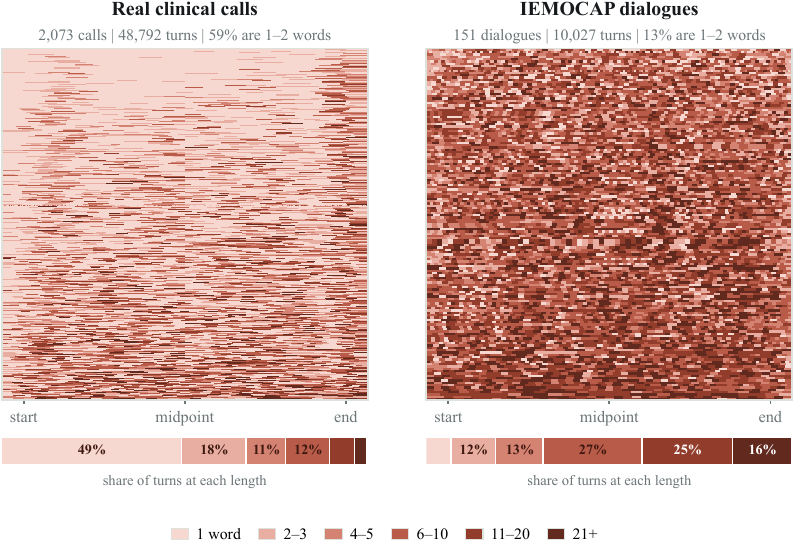}
  \caption{Distribution of word counts across over 2000 clinical conversations. Each row is one conversation, ordered by mean utterance length; the horizontal axis runs from the start of the conversation to its end. Clinical calls are dominated by 1-2 patient turns throughout. The acted corpus contains proportionally under a quarter as many.}
  \label{fig:seq}
\end{figure}

Figure~\ref{fig:seq} shows the distribution of utterance lengths from the beginning to the end of the dialogues for 2073 real clinical calls (patient utterances only) in comparison to the 151 dialogues present in IEMOCAP (utterances of both actors). %

\paragraph{Performance drops on the clinical calls, but telephony bandwidth is not the explanation.}

Models score lower on the clinical calls than on the acted set in all six model-and-input cells. Pooled across the three models the difference is
\SI{-8.1}{pp}, with a \SI{95}{\percent} interval of \SIrange{-15.6}{-0.8}{pp}; pooled across the two frontier models it is \SI{-8.7}{pp} (\SIrange{-16.7}{-0.9}{pp}).
Figure~\ref{fig:corpuscontrast} gives every cell and the pooled row. To test whether the audio-only differences were attributable to recording quality (\SI{16}{\kilo\hertz} for the acted recordings versus \SI{8}{\kilo\hertz} for the real calls), we evaluated the acted corpus' audio-only condition both in its original form and after applying the full telephony chain (\SI{8}{\kilo\hertz}, \SIrange{300}{3400}{\hertz} bandpass, and $\mu$-law encoding). Performance changed only slightly: from \SI{83.3}{\percent} to \SI{82.2}{\percent} for one model and from \SI{80.0}{\percent} to \SI{83.3}{\percent} for the other. As the effect was small and changed direction across models, channel quality is unlikely to explain the performance gap between the datasets. Figure~\ref{fig:tel} in Appendix~\ref{app:tel} presents the full comparison.

\paragraph{The acted corpus, matched on length.} Figure~\ref{fig:seq} illustrates an acted set whose length profile differs sharply from the clinical one, so a performance gap between them could be a length effect rather than a domain effect. The acted \num{150} turn evaluation set extracted from IEMOCAP was therefore built to have a word count distribution more closely matched to that of the real clinical calls; Table~\ref{tab:lengthn} shows the distribution of both evaluation sets.

Table~\ref{tab:matched} shows the best model's (Gemini 2.5 Flash) balanced accuracy by turn length for both evaluation sets when given only audio and only text. An overall corpus gap of about eight points remains after length matching, so the gap between spontaneous and acted speech is not just a matter of turn length. At one word, however, the gap disappears: the two corpora are within four points, and text performs slightly worse on the acted corpus (\SI{51.2}{\percent}) than on the clinical one (\SI{54.8}{\percent}), with recall of \SI{11.1}{\percent} and \SI{9.5}{\percent}, respectively (Table~\ref{tab:oneword} in Appendix~\ref{app:bylength}). The short-turn failure therefore generalises beyond the clinical recordings and patients: it reappears on the benchmark once similarly short turns are present. Above one word, where buckets hold \numrange{10}{34} clinical turns each, the gap varies and favours the acted corpus in seven of the ten audio and text comparisons (Table~\ref{tab:matched}).

\begin{table}[H]
  \caption{Best model's balanced accuracy by turn length, clinical against acted,
    \num{150} turns each with matched length profiles. The clinical set
    balanced \num{75}/\num{75} and the acted set \num{76}/\num{74}. $n$ gives
    the clinical and acted turns in each bucket, and Figure~\ref{fig:cilength}
    gives each bucket's interval.}
  \label{tab:matched}
  \centering
  \begin{tabular}{lccccc}
    \toprule
    & & \multicolumn{2}{c}{Audio} & \multicolumn{2}{c}{Text} \\
    \cmidrule(lr){3-4}\cmidrule(lr){5-6}
    Turn length & $n$ & Clinical & Acted & Clinical & Acted \\
    \midrule
    1 word       &   47 / 41 & \SI{60.0}{\percent} & \SI{62.9}{\percent} & \SI{54.8}{\percent} & \SI{51.2}{\percent} \\
    2--3 words   &   24 / 20 & \SI{71.9}{\percent} & \SI{59.4}{\percent} & \SI{56.2}{\percent} & \SI{68.8}{\percent} \\
    4--5 words   &   10 / 11 & \SI{60.0}{\percent} & \SI{71.7}{\percent} & \SI{90.0}{\percent} & \SI{61.7}{\percent} \\
    6--10 words  &   34 / 28 & \SI{73.5}{\percent} & \SI{85.7}{\percent} & \SI{64.7}{\percent} & \SI{85.7}{\percent} \\
    11--20 words &   22 / 24 & \SI{65.6}{\percent} & \SI{81.4}{\percent} & \SI{70.8}{\percent} & \SI{77.9}{\percent} \\
    21+ words    &   13 / 26 & \SI{80.0}{\percent} & \SI{80.0}{\percent} & \SI{63.7}{\percent} & \SI{77.6}{\percent} \\
    \midrule
    All          & 150 / 150 & \SI{71.3}{\percent} & \SI{79.4}{\percent} & \SI{66.7}{\percent} & \SI{74.8}{\percent} \\
    \bottomrule
  \end{tabular}
\end{table}

\subsection{On short turns, transcript-only detection collapses to chance; audio degrades but does not}
\label{sec:c2}
\label{sec:whytext}

On one-word turns (\num{47} of the \num{150} clinical turns and \num{41} of the
acted), no transcript-only condition has an interval that excludes chance. The strongest reach \SI{54.8}{\percent} balanced accuracy and the weakest \SI{47.1}{\percent}. Recall on those turns falls to between \SI{9.5}{\percent} and \SI{22.2}{\percent} (Table~\ref{tab:oneword}): these systems almost never flag a single-word answer, whatever its delivery. Audio degrades on the same turns without collapsing for some models, reaching \SIrange{42.1}{70.3}{\percent} with recall between \SI{23.8}{\percent} and \SI{66.7}{\percent}, but not uniformly (Table~\ref{tab:oneword}): the floor is the open model on clinical audio, which sits below chance, while Gemini 2.5 Pro on the acted corpus is the only one of the twelve conditions whose interval excludes chance outright. Figure~\ref{fig:oneword} shows all twelve conditions against the chance line.

\paragraph{Predictions track length more than the labels do.}
\label{sec:shortcut}

\begin{figure}[H]
  \centering
  \includegraphics[width=0.7\linewidth]{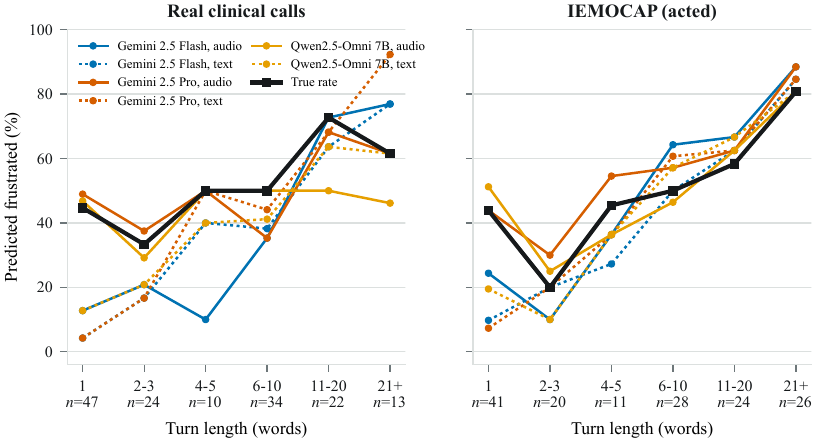}
  \caption{Predicted-frustration rate against turn length for all three models, against the true rate in black. Every transcript condition starts far below the true rate on the shortest turns and climbs more steeply than it.}
  \label{fig:lengthbias}
\end{figure}

The rate at which each condition predicts frustration rises with turn length far more steeply than the true rate does (Figure~\ref{fig:lengthbias}). On the clinical calls the true rate
varies by \SI{39}{pp} across length buckets, whereas frustration prediction rate in the transcript condition varies by a minimum of \SI{51}{pp} with length, and peaks at a variation of \SI{88}{pp}. Two audio conditions vary by \SI{33}{pp} and \SI{21}{pp}, at or below the true variation, so access to delivery partly breaks the dependence on length (Table~\ref{tab:spread}).

Strip delivery from a one-word turn and almost no length signal and almost no lexical signal remain. A logistic regression given only the word count, and never the audio, reaches \SI{58.0}{\percent} on the clinical set (dialogue-disjoint train/test split; Appendix~\ref{app:probetransfer}), and is stable across resamples: \SI{57.9}{\percent} $\pm$ \SI{0.3}{pp} across five held-out-session folds. The evidence here is correlational:
it establishes that predictions track length more strongly than the labels do, not that length is the causal input.

Standard acoustic features cannot be measured on these turns either. We recomputed each of
the \num{88} eGeMAPS parameters~\citep{eyben2016gemaps} on short windows drawn from inside longer turns, so spread
across windows is measurement noise rather than a difference between turns: at
\SI{0.5}{\second}, the median one-word turn, formant-frequency variability reaches an
intraclass correlation of \num{0.17} and no parameter reaches \num{0.70}
(Appendix~\ref{app:measure}). The models are not given these
features, so this does not explain their behaviour, but an interpretable acoustic detector
is not an available substitute at these turn lengths.

\begin{figure}[H]
  \centering
  \includegraphics[width= 0.9 \linewidth]{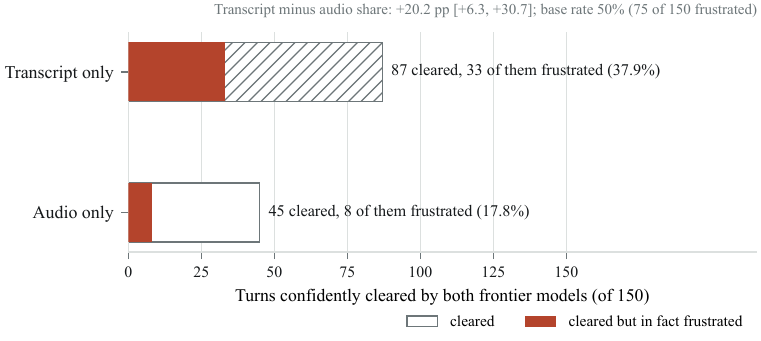}
  \caption{Turns each input modality confidently clears, and how many of them were in fact frustrated.  The transcript clears \num{87} of \num{150} turns and \SI{37.9}{\percent} of those were frustrated; audio clears \num{45} and \SI{17.8}{\percent} were. The audio rate holds only when its prompt directs the model to delivery (Section~\ref{sec:instruction}).}
  \label{fig:clearing}
\end{figure}

\paragraph{The failure is a safety failure.}
\label{sec:safety}
A triage system does not report accuracy; it decides which calls a clinician never sees. A confidently cleared turn is one that both frontier models called not frustrated on every repeat prediction. Of the turns each input modality confidently clears, the transcript condition clears \num{87} and \SI{37.9}{\percent} of them were in fact frustrated. Audio clears \num{45} and \SI{17.8}{\percent} were frustrated (Figure~\ref{fig:clearing}). That lower rate depends on the audio prompt directing the model to delivery: under a prompt identical for both inputs, audio clears \num{81} turns of which \SI{33.3}{\percent} were frustrated, against \num{94} and \SI{38.3}{\percent} for the transcript (Section~\ref{sec:instruction}). These are shares within a sample balanced \num{75}/\num{75} and screened for dissatisfaction, so the absolute levels overstate deployment prevalence; the text-versus-audio contrast, not the absolute rate, is what the comparison supports. The wrongly cleared turns concentrate in the shortest ones: \num{19} of the \num{33} are one or two words, against short turns making up \SI{35}{\percent} of frustrated turns overall (Table~\ref{tab:failures}, Figure~\ref{fig:failures}). Neither the threshold nor a single call carries this: sweeping the clearing threshold leaves text between \SI{37.5}{\percent} and \SI{40.0}{\percent} wrong while audio rises from \SI{17.8}{\percent} to \SI{31.7}{\percent}, and holding out each call in turn moves neither condition far from its full-sample value (Figure~\ref{fig:robust}).

\begin{table}[H]
  \caption{Where the transcript condition's silent failures fall. A silent
    failure is a confidently cleared turn that was labelled as frustrated by the human annotators (a falsely-cleared turn). Short turns are a third of the frustrated turns and well over
    half of the silent failures, and the miss rate falls as turns lengthen.
    Figure~\ref{fig:failures} plots the two shares.}
  \label{tab:failures}
  \centering
  \begin{tabular}{lccccc}
    \toprule
    & \multicolumn{2}{c}{Frustrated turns} & \multicolumn{2}{c}{Silent failures} & \\
    \cmidrule(lr){2-3}\cmidrule(lr){4-5}
    Turn length & $n$ & share & $n$ & share & Missed \\
    \midrule
    1--2 words   & 26 & \SI{34.7}{\percent} & 19 & \SI{57.6}{\percent} & \SI{73.1}{\percent} \\
    3--10 words  & 25 & \SI{33.3}{\percent} & 10 & \SI{30.3}{\percent} & \SI{40.0}{\percent} \\
    11+ words    & 24 & \SI{32.0}{\percent} &  4 & \SI{12.1}{\percent} & \SI{16.7}{\percent} \\
    \midrule
    All          & 75 & \SI{100}{\percent}  & 33 & \SI{100}{\percent}  & \SI{44.0}{\percent} \\
    \bottomrule
  \end{tabular}
\end{table}

\paragraph{Context does not rescue the transcript.} To understand if the text and audio conditions were handicapped by seeing only one turn in isolation, we ran a series of context experiments. We swept the number of preceding patient turns supplied from zero to
six. Text improves by roughly two points and then plateaus: text given six turns of prior conversation still scores below audio
given none (\SI{68.7}{\percent} against \SI{70.7}{\percent}). Audio does not benefit either, and on short turns degrades by \SI{14.7}{pp} across the sweep,
consistently across repeats. Figure~\ref{fig:context} in
Appendix~\ref{app:contextfig} gives the full curves. Giving both inputs the agent's preceding turn, which the patient's turn answers, does not rescue the transcript either: its balanced accuracy changes by between \SI{-2.0}{pp} and \SI{1.3}{pp} across the three models, and recall on frustrated one-word turns rises only from \num{2} to between \num{3} and \num{5} of \num{21}. With the agent's turn, Gemini 2.5 Pro's audio recovers \num{15} of the \num{21}. Appendix~\ref{app:agentturn} gives the prompts, every model and input with and without the agent's turn (Table~\ref{tab:agentturn}), and audio minus transcript in each run (Table~\ref{tab:agentturndiff}).

\subsection{Audio's improvement on transcripts is model dependent}
\label{sec:c3}
\label{sec:complementary}

\begin{figure}[H]
  \centering
  \includegraphics[width=0.8\linewidth]{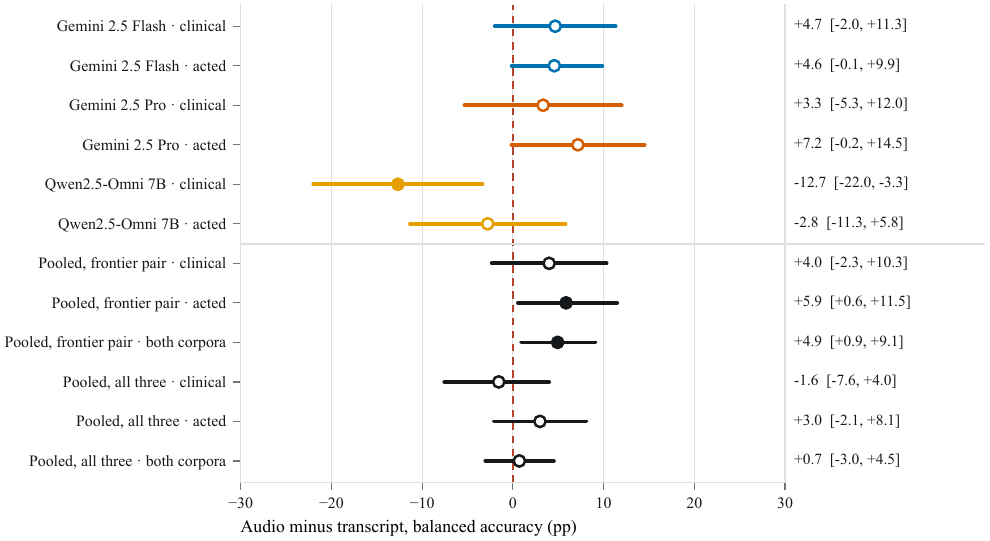}
  \caption{Comparison of audio against the transcript, as differences with \SI{95}{\percent} intervals from a \num{10000}-draw bootstrap over turns. A filled marker marks an interval that excludes zero. No frontier model's audio gain excludes zero on its own. Pooled, the frontier pair's gain excludes zero on the acted corpus and on both corpora together, though not on the real calls alone, and adding the open model removes it. The open model is worse with audio on the real calls, by an interval that excludes zero.}
  \label{fig:modalitycontrast}
\end{figure}

Figure~\ref{fig:modalitycontrast} shows the difference in the models' performance when given audio versus transcripts. Pooled across the two frontier models across both corpora, audio outperformed transcript-only input by \SI{4.9}{pp} (\SIrange{0.9}{9.1}{pp}). On the clinical calls, the individual model estimates were \SI{4.7}{pp} (\SIrange{-2.0}{11.3}{pp}) and \SI{3.3}{pp} (\SIrange{-5.3}{12.0}{pp}); however, neither confidence interval excluded zero at this sample size.

In contrast, the smaller open model performed \SI{12.7}{pp} worse with audio than with transcripts (\SIrange{-22.0}{-3.3}{pp}) on the clinical calls. Including this model in the pooled analysis reduced the overall audio advantage to \SI{0.7}{pp}. Together, these results suggest that the benefit of audio input depends on the model. Figure~\ref{fig:headline} compares all models, input modalities, and corpora on a common scale.

\paragraph{The frontier gain depends on the instruction.} Our audio and transcript prompts differ in what they direct the model to use (Appendix~\ref{app:prompts}). Given one prompt identical for both inputs, the frontier difference disappears. %
Transcript-only detection fails on short turns under every prompt we tried; audio recovers them only when asked to listen for delivery (Section~\ref{sec:instruction}).

\paragraph{The two inputs are complementary.} Because audio and the transcript get different turns right rather than failing together, an oracle that picked the better input for each turn would reach \SI{78.0}{\percent}; a simple vote across the models and inputs reaches only \SI{66.0}{\percent}, below either frontier model's audio alone, so the complementarity is real but a naive combination does not capture it (Appendix~\ref{app:compl}).

\subsection{The findings hold with more calls and more models}
\label{sec:thirty}

\begin{figure}[H]
  \centering
  \includegraphics[width=0.9\linewidth]{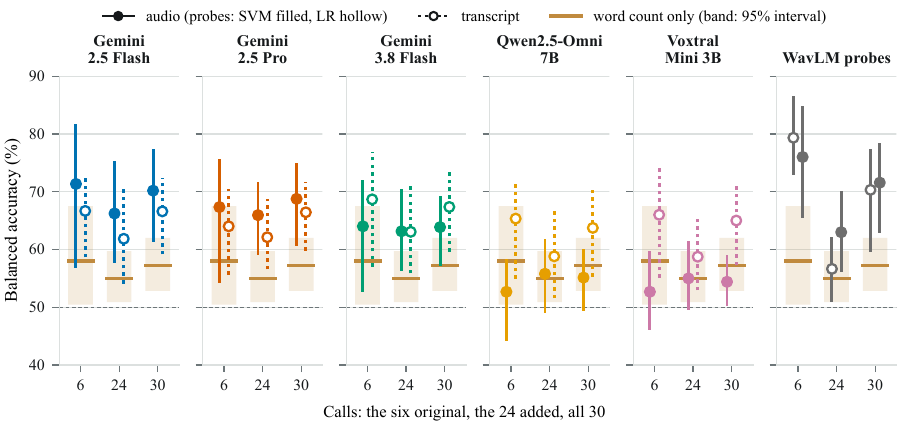}
  \caption{Balanced accuracy on the six original calls, the 24 added calls and all 30, with \SI{95}{\percent} bootstrap intervals over calls. Audio is solid and filled, the transcript dotted and hollow. The last panel shows the two WavLM probes trained on the 30 calls (Section~\ref{sec:probes}). The bar and band mark the word-count-only baseline and its interval.}
  \label{fig:thirtyheadline}
\end{figure}

We repeat every comparison within the clinical calls made without context on all \num{714} turns of the 30 calls (Section~\ref{sec:dataset}), and add two models: Gemini 3.8 Flash, the newer model in the frontier pair's family, and Voxtral Mini 3B, a second small open model. Figure~\ref{fig:thirtyheadline} shows balanced accuracy for each model and input on the six, the 24 and the 30 calls, and Appendix~\ref{app:thirty} gives the remaining measures.

\begin{figure}[H]
  \centering
  \includegraphics[width=0.9\linewidth]{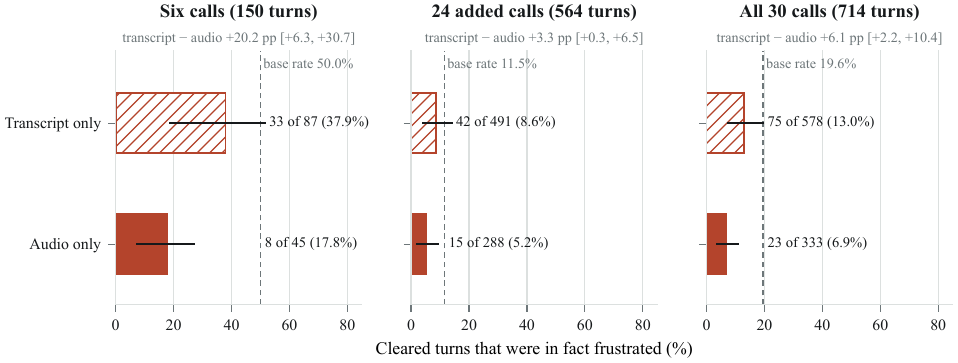}
  \caption{Share of the frontier pair's confidently cleared turns that were frustrated, on the six calls (\num{33} of \num{87} for the transcript, \num{8} of \num{45} for audio), the 24 (\num{42} of \num{491}, \num{15} of \num{288}) and all 30 (\num{75} of \num{578}, \num{23} of \num{333}), with \SI{95}{\percent} bootstrap intervals over calls. Each part's frustrated base rate is marked: \SI{50}{\percent}, \SI{11.5}{\percent} and \SI{19.6}{\percent}.}
  \label{fig:thirtyclearing}
\end{figure}

The results on 30 calls follow those on six. Every model's transcript flags \num{3} or \num{4} of the \num{39} frustrated one-word turns (Figure~\ref{fig:thirtyoneword} in Appendix~\ref{app:thirty}). The frontier pair's transcript wrongly clears more frustrated turns than its audio: \SI{13.0}{\percent} of \num{578} cleared turns against \SI{6.9}{\percent} of \num{333}, a difference of \SI{6.1}{pp} (\SIrange{2.2}{10.4}{pp}; Figure~\ref{fig:thirtyclearing}). Its audio gain stays positive, at \SI{3.0}{pp} (\SIrange{-1.0}{7.5}{pp}). Qwen2.5-Omni is again worse with audio, by \SI{8.6}{pp} (\SIrange{-14.1}{-1.6}{pp}).

The two added models split. Voxtral behaves like Qwen2.5-Omni and is worse with audio, by \SI{10.6}{pp} (\SIrange{-16.5}{-3.6}{pp}). Gemini 3.8 Flash reverses the pattern of the 2.5 models in its own family: it is \SI{3.5}{pp} worse with audio (\SIrange{-6.0}{0.0}{pp}), and from audio it flags \num{4} of the \num{39} frustrated one-word turns, against \num{9} and \num{23} for the 2.5 models. Gemini 2.5 Pro's \num{23} come with \num{77} false alarms among the \num{278} one-word turns that were not frustrated.

\begin{figure}[H]
  \centering
  \includegraphics[width=\linewidth]{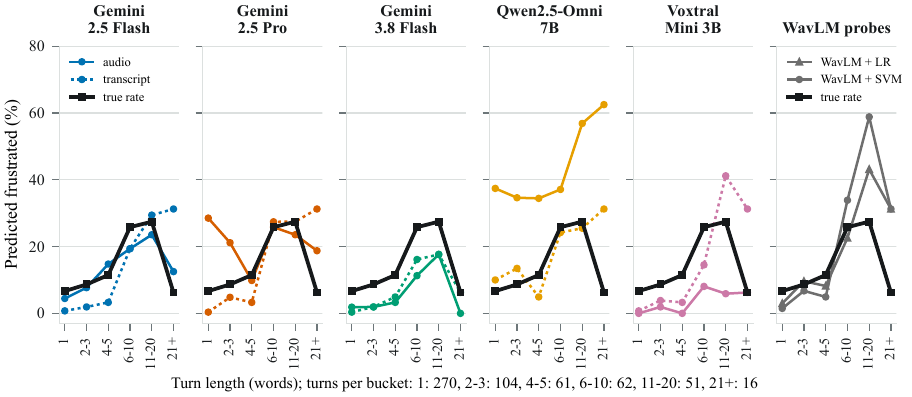}
  \caption{Predicted-frustration rate by turn length on the 24 added calls (\num{564} turns), against the true rate in black, for the five models and the two WavLM probes trained on the 30 calls. Audio is solid, the transcript dotted.}
  \label{fig:thirtylengthbias}
\end{figure}

On the 24 added calls, whose turns keep their natural prevalence, predictions again track length more than the labels do (Figure~\ref{fig:thirtylengthbias}). The true rate varies by \SI{21.2}{pp} across length buckets. The transcript's predicted rate varies by \SIrange{26.3}{40.4}{pp} for four of the five models, and by \SI{17.3}{pp} for Gemini 3.8 Flash. On one-word turns, \SI{6.7}{\percent} of which are frustrated, every transcript but Qwen2.5-Omni's predicts frustration on at most \SI{0.7}{\percent}; Gemini 2.5 Pro's audio predicts it on \SI{28.5}{\percent}.

\subsection{The instruction decides whether audio helps}
\label{sec:instruction}

Our audio and transcript prompts differ in what they direct the model to use (Appendix~\ref{app:prompts}), so a difference between the inputs could come from the instruction rather than the input. We therefore reran all five models on all 30 calls with one prompt identical for both inputs (Table~\ref{tab:neutralfive}; Figure~\ref{fig:promptgap} in Appendix~\ref{app:thirty} plots the differences).

On the six calls, the identical prompt makes the frontier gain disappear: \SI{-1.3}{pp} (\SIrange{-7.7}{4.7}{pp}) for Gemini 2.5 Flash and \SI{0.7}{pp} (\SIrange{-7.0}{6.8}{pp}) for Gemini 2.5 Pro. Their audio flags \num{0} and \num{1} of the \num{21} frustrated one-word turns, against \num{5} and \num{13} when asked to listen, and the transcript's \num{2}. The transcript barely moves. Frontier clearing converges the same way: audio clears \num{81} turns of which \SI{33.3}{\percent} were frustrated, against \num{94} turns and \SI{38.3}{\percent} for the transcript. Gemini 3.8 Flash moves from \SI{-4.7}{pp} (\SIrange{-8.7}{-0.5}{pp}) to \SI{-0.7}{pp} (\SIrange{-5.1}{2.6}{pp}). The open models diverge. Qwen2.5-Omni's deficit does not depend on the prompt (\SI{-12.7}{pp} and \SI{-12.0}{pp}). Voxtral's shrinks from \SI{-13.3}{pp} (\SIrange{-21.2}{-2.8}{pp}) to \SI{-3.3}{pp} (\SIrange{-11.6}{7.2}{pp}), because its audio rises from \SI{52.7}{\percent} to \SI{62.0}{\percent}.

The 30 calls show the same pattern. Under the identical prompt, no model's audio--text interval excludes zero: \SI{-0.1}{pp}, \SI{0.9}{pp}, \SI{-0.7}{pp}, \SI{-5.7}{pp} and \SI{-1.2}{pp} for Gemini 2.5 Flash, 2.5 Pro, 3.8 Flash, Qwen and Voxtral. Pro's audio flags \num{4} of the \num{39} frustrated one-word turns in place of \num{23}. Audio's clearing advantage mostly disappears: \SI{12.0}{\percent} of the \num{501} turns audio clears were frustrated, against \SI{13.2}{\percent} for the transcript.

\begin{table}[htbp]
  \caption{Input-specific prompts against one prompt identical for both inputs, on the six calls, for all five models. Balanced accuracy, the paired audio-minus-text difference in percentage points with its \SI{95}{\percent} bootstrap interval over calls, and recall on the \num{21} frustrated one-word turns.}
  \label{tab:neutralfive}
  \centering
  \small
  \begin{tabular}{llcccc}
    \toprule
    & & & & & One-word recall \\
    Model & Prompts & Audio & Text & Audio $-$ text & audio / text \\
    \midrule
    Gemini 2.5 Flash & Input-specific & 71.3 & 66.7 & $+4.7$ [$-2.6$, $+10.9$] & 5 / 2 \\
                     & Identical      & 66.0 & 67.3 & $-1.3$ [$-7.7$, $+4.7$]  & 0 / 2 \\
    \addlinespace[3pt]
    Gemini 2.5 Pro   & Input-specific & 67.3 & 64.0 & $+3.3$ [$-6.2$, $+11.9$] & 13 / 2 \\
                     & Identical      & 67.3 & 66.7 & $+0.7$ [$-7.0$, $+6.8$]  & 1 / 2 \\
    \addlinespace[3pt]
    Gemini 3.8 Flash & Input-specific & 64.0 & 68.7 & $-4.7$ [$-8.7$, $-0.5$]  & 1 / 2 \\
                     & Identical      & 66.7 & 67.3 & $-0.7$ [$-5.1$, $+2.6$]  & 0 / 2 \\
    \addlinespace[3pt]
    Qwen2.5-Omni 7B  & Input-specific & 52.7 & 65.3 & $-12.7$ [$-19.5$, $-7.0$] & 8 / 2 \\
                     & Identical      & 56.7 & 68.7 & $-12.0$ [$-24.3$, $-1.5$] & 2 / 3 \\
    \addlinespace[3pt]
    Voxtral Mini 3B  & Input-specific & 52.7 & 66.0 & $-13.3$ [$-21.2$, $-2.8$] & 0 / 2 \\
                     & Identical      & 62.0 & 65.3 & $-3.3$ [$-11.6$, $+7.2$]  & 1 / 8 \\
    \bottomrule
  \end{tabular}
\end{table}

\subsection{Trained acoustic probes}
\label{sec:probes}

A probe on frozen speech representations needs no instruction to attend to delivery. We score two kinds on the 30 calls (Figure~\ref{fig:probesthirty}). The first is trained only on acted speech, the probe of Appendix~\ref{app:probes}. The second is trained on the clinical calls and scored out of fold (Section~\ref{sec:setup}).

The acted-trained probe does not carry to the added calls. On the six calls the band-matched WavLM SVM reaches \SI{74.0}{\percent} and flags \num{14.0} of the \num{21} frustrated one-word turns on average across folds. On the 24 added calls it reaches \SI{62.8}{\percent} (\SIrange{57.5}{69.5}{\percent}) and flags \num{7.8} of \num{18}. With its threshold chosen on held-out clinical calls, it flags \num{3.8} of \num{18}.

Training on clinical calls gives the best overall accuracy, but not on short turns. Trained on the 30 calls, the SVM reaches \SI{71.6}{\percent} (\SIrange{63.2}{78.2}{\percent}) and the logistic regression \SI{70.3}{\percent} (\SIrange{59.8}{77.1}{\percent}), level with Gemini 2.5 Flash on audio (\SI{70.2}{\percent}) and well above word count. Most of that comes from the six calls (\SI{76.0}{\percent} and \SI{79.3}{\percent}); on the 24 they reach \SI{63.0}{\percent} and \SI{56.6}{\percent}. The logistic regression flags \num{15} of the \num{21} frustrated one-word turns in the six calls and none of the \num{18} in the 24; the SVM flags \num{8} and none. Of the turns the SVM confidently clears across its three split seeds, \SI{10.2}{\percent} were frustrated, between the frontier pair's audio (\SI{6.9}{\percent}) and transcript (\SI{13.0}{\percent}). The trained probes get long turns right and, on the added calls, clear the short ones much as the transcript does.

The six calls are a balanced, triple-annotated sample and the 24 a single-annotated sample at \SI{11.5}{\percent} prevalence, so the drop from the six to the 24 mixes call difficulty, label source and class balance.

\begin{figure}[H]
  \centering
  \includegraphics[width=0.9\linewidth]{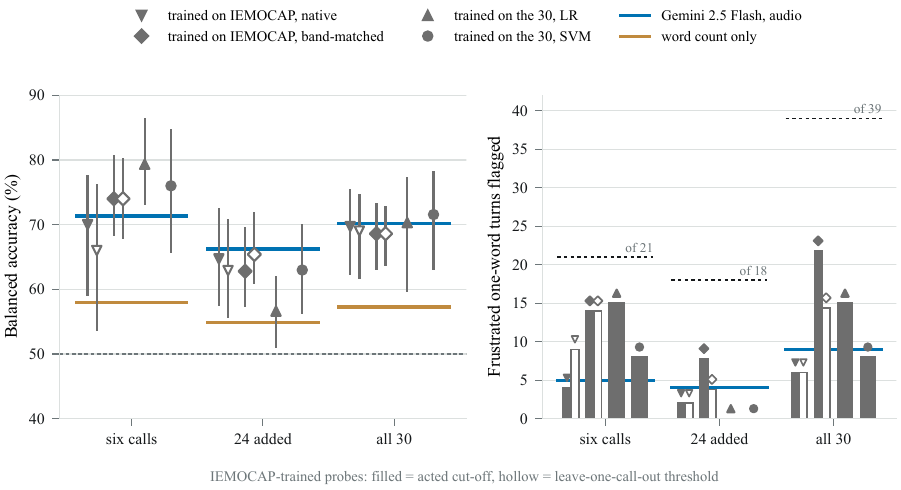}
  \caption{The four WavLM probes on the six, the 24 and all 30 calls: trained on IEMOCAP with native and band-matched audio, scored at the acted cut-off (filled) and with a threshold chosen leaving one call out (hollow), and trained on the 30 calls with a logistic regression and an SVM. Left: balanced accuracy with \SI{95}{\percent} bootstrap intervals over calls. Right: frustrated one-word turns flagged, of \num{21}, \num{18} and \num{39}; the band-matched counts average five folds. Reference lines: Gemini 2.5 Flash audio and word count only.}
  \label{fig:probesthirty}
\end{figure}

\section{Discussion}
\label{sec:discussion}

\paragraph{A transcript-only system clears short frustrated turns silently, and with high confidence.} On a short turn there is almost nothing to read, and the absence of explicit textual evidence reads as calm, so the miss is silent: the turn is cleared with nothing to flag it. Accuracy is therefore the wrong measure. Two systems can tie on balanced accuracy yet clear entirely different turns: text clears far more than audio and, on the six calls, is wrong about more than a third of them, and those are the short ones, often at a call's onset where an escalation still matters. The starkest case is a patient answering ``No.'' with opposite states minutes apart (Appendix~\ref{app:escalation}): identical words, and text clears both.

\paragraph{The failure tracks turn length, not our patients or recordings.} It reappears on the acted benchmark as soon as that benchmark is made to contain short turns, which argues against our dataset, telephone channel, or clinical setting as the explanation, though length remains confounded with response type and dialogue position in both corpora. We would expect the same wherever a transcript-based system must read affect from brief spoken answers, but we show it only for frustration on clinical calls.

\paragraph{Standard emotion benchmarks such as IEMOCAP cannot show that transcript-only detection fails on short turns, because they contain almost none.} An evaluation run on them dilutes the failure until it looks like a pass, which is more dangerous than a test that plainly does not apply, because it still reads as validation. The fix is cheap: report the length distribution of the turns evaluated, test on the lengths a system will meet in deployment, and add a length-only baseline. On our data word count alone reaches \SIrange{57}{58}{\percent}; a model that cannot beat word count has not shown it is using the voice.

\paragraph{Adding audio is not automatically an upgrade; the model and its instruction decide whether it helps.} Directed to delivery, it raises the two Gemini 2.5 models by about \SI{4}{pp} on the six calls and \SI{3}{pp} across 30, estimates whose intervals include zero. Under a prompt identical for both inputs, the gain is gone for every model we tested. The newer model in the same family and both small open models do worse with audio than without. These models receive the signal and often do not act on it, so the limit lies in using delivery, and it varies by model and by prompt. An audio system should be measured on the target distribution, under the prompt it will run with, before it is assumed better. That audio adds over the transcript for clinical affect is consistent with prior work on acted distress~\citep{hu2026cbtaudio}. Our contribution is not that audio helps. It is that on real deployed speech, whatever margin audio offers sits in the one- to three-word turns that a transcript silently clears, and is realised only when the model is directed to use it.

\paragraph{For deployment, a short patient turn should not be cleared on the transcript alone; it is a candidate for routing to an audio model or a person, subject to the review load that creates.} No system we tested makes that route safe on its own. Directed audio from the strongest model recovers most frustrated one-word turns, but with many false alarms. Smaller and newer models recover few. Trained probes recover them on some calls and not on others (Section~\ref{sec:probes}). Our claim is about the measurement and the regime, not a detector ready to ship.

\paragraph{A successor model is not a drop-in replacement for listening.} The two Gemini 2.5 models that gain from audio are scheduled for retirement. Their successor, Gemini 3.8 Flash, reads transcripts as well as Gemini 2.5 Flash (\SI{67.4}{\percent} against \SI{66.6}{\percent} on 30 calls) but does worse from audio, and catches \num{4} of \num{39} frustrated one-word turns where 2.5 Flash catches \num{9} and 2.5 Pro \num{23}. A system that relies on audio for short turns has to be re-measured whenever its model changes, because the property it relies on is not preserved across versions.

\paragraph{Training on clinical speech improves the average without solving short turns.} A WavLM probe trained on the 30 calls matches the best prompted model overall, and it needs no instruction. But its one-word success on the six calls does not reappear on the 24 added calls, where it flags none of the \num{18} frustrated one-word turns. The six calls differ from the 24 in label source, class balance and prevalence, so we cannot attribute the drop to any one of them. A trained probe, too, has to be shown to catch short frustrated turns on the calls it will see, not only to score well overall.

\paragraph{Limitations.} The data come from a single deployed agent on one clinical pathway, in one language and healthcare system, and judge a single affective state as a binary. The 24 added calls were labelled by Annotator~1 alone, so their labels carry no agreement estimate, and their transcripts are uncorrected Whisper output~\citep{radford2023whisper}. The 30-call set combines a balanced sample from the six calls with every labelled turn of the 24, so its pooled prevalence describes neither, and we rest prevalence-sensitive claims on the 24. The acted control, telephony, context, agent-turn, minimal-pair and onset experiments cover only the six calls and the three original models. We evaluate five models zero-shot, three from one provider, and train only frozen-encoder probes. The Gemini 2.5 models are scheduled for retirement. Appendix~\ref{app:limitations} expands these.

\section{Conclusion} 
\label{sec:conclusion}
A deployed clinical voice agent decides what to escalate from the transcript, not the voice, and that choice fails precisely where clinical speech lives. Patient turns are short: roughly \SI{59}{\percent} are one or two words. On those turns transcript-only frustration detection stops working. It does not read them wrong so much as decline to judge them, which in a binary decision reads as calm. This holds on six triple-annotated calls and on 30, for each of five models, which flag \num{3} or \num{4} of the \num{39} frustrated one-word turns, the same three for every model. The cost is a safety one rather than an accuracy one, because a triage system decides which calls a clinician never sees. Across 30 calls, \SI{13.0}{\percent} of the turns the transcript confidently clears were frustrated, against \SI{6.9}{\percent} for audio directed to delivery. Audio is no fix in itself. Directed to delivery, it helps two frontier models by about three points; without that instruction the gain disappears, and a newer model and two smaller ones do worse with audio under the directed prompt. A probe trained on the clinical calls matches the best prompted model overall, yet flags none of the \num{18} frustrated one-word turns in the 24 added calls. The failure is much less visible on acted corpora, which are dominated by longer turns, and reappears when the acted set is made short. The consequence is narrow and firm: an evaluation that does not report the length distribution of its turns cannot show that a transcript-based system will hold up in deployment, and a short patient turn should not be cleared on the transcript alone.

\section*{Ethics statement}
The real-call corpus is 30 recordings from a deployed clinical voice agent, collected during routine service delivery. Patients consented at the point of collection to the use of their data for research and service improvement. The identity-verification section of each call was removed from the audio and transcript before analysis. Real-call audio was processed only on local machines and through Vertex AI under a data processing agreement with Google. No real-call audio, transcripts or patient speech are released; the paper reports aggregate results only. IEMOCAP is a corpus of acted dyadic conversations, used under its licence agreement.

\bibliographystyle{plainnat}
\bibliography{references}

\appendix
\newpage

\section{Extended limitations}
\label{app:limitations}

\paragraph{Scale and scope.} The six-call evaluation set is \num{150} turns, and the 30-call set \num{714} turns, all from a single clinical pathway, in one language and one healthcare system. Per-call variance is substantial in both. The onset analysis rests on six items and is reported as an illustration.

\paragraph{Selection.} The calls were originally screened using indicators of dissatisfaction, including low feedback scores, so the corpus may over-represent one kind of frustration. Reported frustration rates should not be read as population estimates. The within-corpus comparisons that carry our claims are unaffected, since every condition faces the same data.

\paragraph{Annotation ceiling.} Human agreement bounds what any model result can demonstrate. With $\kappa$ between 0.6 and 0.8, model performance should be read against that ceiling rather than against perfect agreement.

\paragraph{Model coverage.} We evaluate five models zero-shot; three share a provider family. We fine-tune no acoustic model end to end, reporting only frozen-encoder probes (Section~\ref{sec:probes}, Appendix~\ref{app:probes}). Gemini 2.5 Flash and Pro are scheduled for deprecation, and their successor in our evaluation, Gemini 3.8 Flash, uses audio differently (Section~\ref{sec:thirty}).

\paragraph{Transcript quality.} The text condition is only as good as the transcript it receives. On the six calls, every turn's transcript is either agreed by two independent ASR systems or adjudicated by a human against the audio, and the onset examined in Appendix~\ref{app:onset} was verified by listening because the production transcript for it is wrong. The 24 added calls use Whisper large-v3 transcripts. On a one-word turn a single substituted word changes the affective reading entirely.

\paragraph{Annotation of the added calls.} The 24 added calls were labelled by Annotator~1 alone, so they carry no agreement estimate; the six triple-annotated calls give the agreement reference.

\section{Method and configuration}

\subsection{Experimental configuration}
\label{app:config}

\begin{figure}[htbp]
  \centering
  \includegraphics[width=\linewidth]{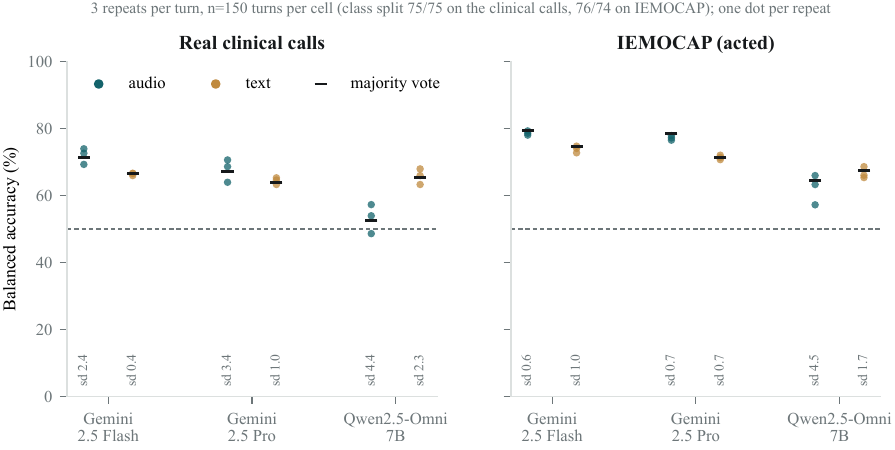}
  \caption{Balanced accuracy of each of the three runs, scored on its own against the human labels; the marker shows the majority vote of the three. Standard deviations range from \SIrange{0.4}{4.4}{pp}.}
  \label{fig:cirepeats}
\end{figure}

Each (model, modality, turn) combination is run three times and the turn's
prediction is the majority vote of the three. The three runs differ in
balanced accuracy by a standard deviation of \SIrange{0.4}{4.4}{pp} on the six
calls and \SIrange{0.1}{2.6}{pp} on the 30, across the five models, and span at
most \SI{8.7}{pp}, for the open model on audio on the six calls (Figure~\ref{fig:cirepeats}), and in
places this exceeds the effect sizes under study, so every headline number is a
majority vote over repeats and every directional claim is additionally checked
per repeat. Redrawing the repeats inside the bootstrap gives the acted frontier
pair's audio advantage of \SI{5.9}{pp} an interval of \SIrange{-0.1}{10.9}{pp}
rather than \SIrange{0.6}{11.5}{pp}; the frontier pair pooled across both
corpora stays above zero at \SIrange{0.6}{9.1}{pp}.
\subsection{How each model is run}
\label{app:serving}

Table~\ref{tab:serving} gives how each model is accessed and called. Every audio request sends the turn's clip as a WAV file cut from the call recording, without resampling. The three Gemini models are called through the Google Gen AI SDK on Vertex AI with a JSON response schema. The two open models run locally. Qwen2.5-Omni runs in llama.cpp's \texttt{llama-server} through its OpenAI-compatible chat endpoint, with the clip sent inline as base64 and a JSON schema on the response. Voxtral runs in Hugging Face Transformers with no response schema; the prompt ends with a request for the same JSON object, and the reply is parsed from its first JSON block, falling back to the label word. API and server errors get up to four attempts with exponential backoff; a turn whose reply cannot be parsed counts as a failed request.

\begin{table}[htbp]
  \caption{How each model is run.}
  \label{tab:serving}
  \centering
  \footnotesize
  \begin{tabular}{lllll}
    \toprule
    Model & Identifier & Access & Weights & Sampling \\
    \midrule
    Gemini 2.5 Flash & \texttt{gemini-2.5-flash} & Vertex AI & -- & temperature 1.0 \\
    Gemini 2.5 Pro & \texttt{gemini-2.5-pro} & Vertex AI & -- & temperature 1.0 \\
    Gemini 3.8 Flash & \texttt{gemini-3.8-flash} & Vertex AI & -- & temperature 1.0 \\
    Qwen2.5-Omni 7B & \texttt{ggml-org/Qwen2.5-Omni-7B-GGUF} & llama.cpp & Q8\_0 & temperature 1.0 \\
    Voxtral Mini 3B & \texttt{mistralai/Voxtral-Mini-3B-2507} & Transformers & bf16 & temperature 1.0 \\
    \bottomrule
  \end{tabular}
\end{table}

\subsection{Prompts}
\label{app:prompts}

The prompts were held as consistent as possible across modalities, and differ in what they direct the model to use. Each is given exactly as sent to the model.

\begin{tcolorbox}[
    title=\textbf{Audio prompt},
    colback=blue!3,
    colframe=blue!60!black,
    boxrule=0.7pt,
    arc=2pt,
    breakable
]
Listen to this short clip of a patient speaking on a phone check-in call. Classify the patient's emotional state in THIS clip as exactly one of: frustrated, not\_frustrated. Then explain your reasoning based on what you hear in the audio (tone, pacing, pauses, pitch, etc.) -{}- not the words alone.
\end{tcolorbox}

\begin{tcolorbox}[
    title=\textbf{Transcript-only prompt},
    colback=orange!3,
    colframe=orange!70!black,
    boxrule=0.7pt,
    arc=2pt,
    breakable
]
Below is the transcript of one turn a patient spoke during a phone check-in call. You do not have access to audio -{}- classify the patient's emotional state in THIS turn based on the WORDS ALONE, as exactly one of: frustrated, not\_frustrated. Then explain your reasoning based on the wording, phrasing, and content of what was said.\par\medskip
\texttt{Transcript for this turn: "\{text\}"}
\end{tcolorbox}

The control of Section~\ref{sec:instruction} gives both inputs one prompt, identical word for word, followed by the audio clip or by the line \texttt{Transcript: "\{text\}"}, with the turn's text in place of \texttt{\{text\}}.

\begin{tcolorbox}[
    title=\textbf{Identical prompt, both inputs},
    colback=black!2,
    colframe=black!60,
    boxrule=0.7pt,
    arc=2pt,
    breakable
]
Below is one turn a patient spoke during a phone check-in call. Classify the patient's emotional state in this turn as exactly one of: frustrated, not\_frustrated. Then explain your reasoning.
\end{tcolorbox}

The agent-turn runs of Appendix~\ref{app:agentturn} extend the main prompts with what the agent said immediately before the turn, given in the same words to both inputs, with the agent's words in place of \texttt{\{agent turn\}}.

\begin{tcolorbox}[
    title=\textbf{Audio prompt with the agent's preceding turn},
    colback=blue!3,
    colframe=blue!60!black,
    boxrule=0.7pt,
    arc=2pt,
    breakable
]
Listen to this short clip of a patient speaking on a phone check-in call. Immediately before this clip, the automated assistant said: \texttt{"\{agent turn\}"}\par\medskip
The patient's clip is their reply to that. Use it only to understand what they are responding to -{}- for instance whether they are being asked something for a second time, or being asked something they have already answered. Judge the emotional state from how the PATIENT sounds in their own clip (tone, pacing, pauses, pitch), not from what the assistant said and not from the words alone. Classify the patient's emotional state in THIS clip as exactly one of: frustrated, not\_frustrated. Then explain your reasoning.
\end{tcolorbox}

\begin{tcolorbox}[
    title=\textbf{Transcript-only prompt with the agent's preceding turn},
    colback=orange!3,
    colframe=orange!70!black,
    boxrule=0.7pt,
    arc=2pt,
    breakable
]
Below is the transcript of one turn a patient spoke during a phone check-in call. Immediately before this turn, the automated assistant said: \texttt{"\{agent turn\}"}\par\medskip
The patient's turn is their reply to that. Use it only to understand what they are responding to -{}- for instance whether they are being asked something for a second time, or being asked something they have already answered. You do not have access to audio -{}- judge the emotional state from the WORDING of the patient's own turn, not from what the assistant said. Classify the patient's emotional state in THIS turn as exactly one of: frustrated, not\_frustrated. Then explain your reasoning.\par\medskip
\texttt{Transcript for this turn: "\{text\}"}
\end{tcolorbox}

\subsection{Annotators}
\label{app:annotators}

All \num{224} patient turns in the six calls were labelled by each of the
three annotators, and \num{58} of the \num{672} judgements
(\SI{8.6}{\percent}) were inconclusive. Table~\ref{tab:inconclusive} gives how often a turn was marked
inconclusive and how many turns this removed before the balanced evaluation set
was drawn from the \num{75} frustrated and \num{130} not-frustrated turns that
remained.

\begin{table}[htbp]
  \caption{Inconclusive judgements on the six clinical calls, each patient
    turn labelled by all three annotators.}
  \label{tab:inconclusive}
  \centering
  \small
  \begin{tabular}{lr}
    \toprule
    & Turns \\
    \midrule
    \multicolumn{2}{@{}l}{\textit{Inconclusive judgements per turn}} \\
    None            & 183 \\
    One             &  28 \\
    Two             &   9 \\
    Three           &   4 \\
    \midrule
    \multicolumn{2}{@{}l}{\textit{Final label}} \\
    Frustrated      &  75 \\
    Not frustrated  & 130 \\
    Inconclusive (majority), excluded   & 13 \\
    No majority, excluded               &  6 \\
    \midrule
    All turns       & 224 \\
    \bottomrule
  \end{tabular}
\end{table}

\section{Supporting the corpus comparison}
\begin{table}[htbp]
  \caption{The \num{150} clinical evaluation turns against the whole of
    IEMOCAP (all \num{10039} utterances except the \num{59} with no words, which are breathing or laughter only: \num{9980} turns), on the same measures. The two
    are different regimes, not two samples of one.}
  \label{tab:regime}
  \centering
  \begin{tabular}{lcc}
    \toprule
    & Clinical, \num{150} turns & IEMOCAP, \num{9980} turns \\
    \midrule
    Median turn length            & 4 words & 9 words \\
    Turns of 3 words or fewer     & \SI{47.3}{\percent} & \SI{18.7}{\percent} \\
    Single-word turns             & \SI{31.3}{\percent} & \SI{6.5}{\percent} \\
    A bare yes / no / okay        & \SI{20.0}{\percent} & \SI{1.4}{\percent} \\
    Frustrated turns of 3 words or fewer & \SI{38.7}{\percent} & \SI{15.0}{\percent} \\
    \bottomrule
  \end{tabular}
\end{table}

All three models score lower on the clinical calls than on the acted corpus in every condition.
\begin{figure}[htbp]
  \centering
  \includegraphics[width=0.92\linewidth]{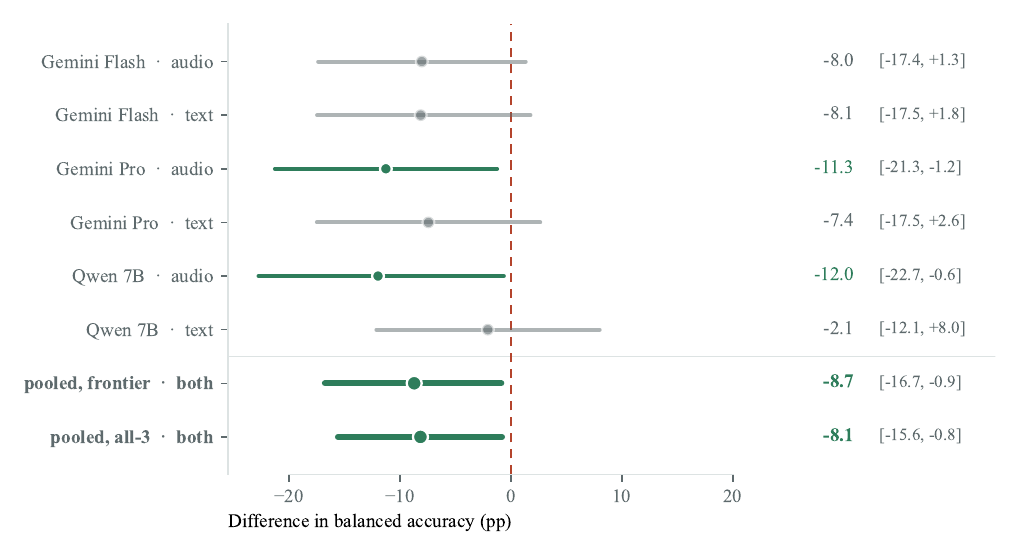}
  \caption{Comparison of real against acted, as differences with \SI{95}{\percent} intervals. Left of zero means the real calls are harder.  Rows below the rule pool across models and are the primary comparison; a green interval excludes zero.}
  \label{fig:corpuscontrast}
\end{figure}

\subsection{Benchmark comparison and the length-matching requirement}
\label{app:benchmark}

A natural comparison is against an established acted emotion corpus. We found
that a naive class-balanced sample from such a corpus can introduce a
length--label confound: in one such sample, frustrated utterances were markedly
longer than neutral ones, to the point that a classifier using \emph{only} word
count---no audio, no lexical content, no model---scored well above chance and
above some evaluated models. On our clinical corpus the same trick reaches
\SI{58.0}{\percent} (Section~\ref{sec:shortcut}). The effect is far weaker in the source corpus as a whole than in that
sample, so it is an artefact of sampling rather than a property of the benchmark.

We therefore hold that any cross-corpus comparison in this setting must match
length distributions explicitly and balance classes \emph{within} length strata.

\subsection{Telephony degradation}
The drop in performance from acted to clinical data cannot be attributed to the degradation in audio quality. Both arms are scored on \num{90} of the \num{150} acted turns, \num{45} frustrated and \num{45} not frustrated.
\label{app:tel}

\begin{figure}[htbp]
  \centering
  \includegraphics[width=0.62\linewidth]{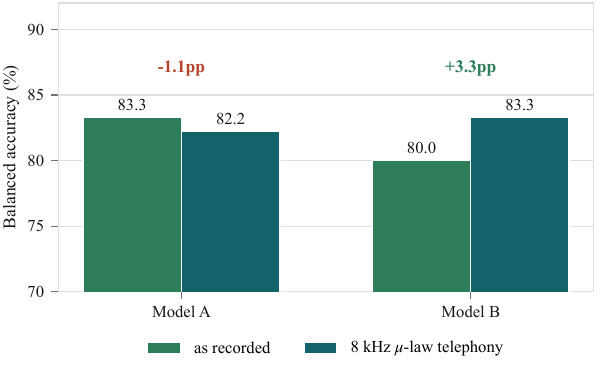}
  \caption{The acted corpus as recorded and through the full telephony chain,
    on \num{90} of the \num{150} acted turns (\num{45} frustrated,
    \num{45} not frustrated). The effect is small and changes sign between models, so the
    clinical--acted gap cannot be attributed to channel bandwidth.}
  \label{fig:tel}
\end{figure}

\subsection{Per-call variance}
\label{app:percall}

Three-way annotator unanimity ranges from \SI{60.9}{\percent} to
\SI{88.1}{\percent} across the six calls, a spread wider
than the differences between modalities. Accuracy varies more across calls than
across conditions; Table~\ref{tab:callci} gives intervals that resample whole calls. The ranking of calls by annotator agreement does not match the
ranking by model accuracy, which argues against the interpretation that model
errors simply track intrinsic ambiguity in the turns.

Turns from one call are not independent, so intervals that resample turns can be
too narrow. Table~\ref{tab:callci} resamples whole calls instead, drawing six
calls with replacement \num{10000} times. The intervals widen, most for one-word
recall, but no difference changes sign. Holding out each call in turn tells the
same story: the frontier audio advantage stays between \SI{2.4}{pp} and
\SI{6.3}{pp} and the open model's deficit between \SI{-9.9}{pp} and
\SI{-14.9}{pp}. One call holds \num{12} of the \num{21} frustrated one-word
turns. Without it, every transcript condition flags none of the remaining
\num{9} and audio flags one or two, so the one-word recall contrast rests mostly
on that call, while the balanced-accuracy contrasts do not.

\begin{table}[htbp]
  \caption{Intervals that resample turns against intervals that resample whole
    calls, on the clinical set. Differences are audio minus text in percentage
    points; the clearance row is the transcript's share of cleared turns that
    were frustrated minus audio's, for the frontier pair. \SI{95}{\percent}
    bootstrap intervals, \num{10000} draws each.}
  \label{tab:callci}
  \centering
  \small
  \begin{tabular}{lccc}
    \toprule
    & Estimate & Over turns & Over calls \\
    \midrule
    Audio $-$ text, Gemini 2.5 Flash & $+4.7$  & [$-2.0$, $+11.3$] & [$-2.6$, $+10.9$] \\
    Audio $-$ text, Gemini 2.5 Pro   & $+3.3$  & [$-5.3$, $+12.0$] & [$-6.2$, $+11.9$] \\
    Audio $-$ text, Qwen2.5-Omni 7B  & $-12.7$ & [$-22.0$, $-3.3$] & [$-19.5$, $-7.0$] \\
    Audio $-$ text, frontier pair    & $+4.0$  & [$-2.3$, $+10.3$] & [$-2.4$, $+9.3$] \\
    \addlinespace[3pt]
    Cleared but frustrated, text $-$ audio & $+20.2$ & --- & [$+6.3$, $+30.7$] \\
    One-word recall, Gemini 2.5 Pro audio  & 61.9    & --- & [0.0, 84.6] \\
    \bottomrule
  \end{tabular}
\end{table}

\section{Supporting the short-turn results}

On one-word turns, audio beats the transcript for every model and corpus except Qwen2.5-Omni 7B on the clinical calls.

\begin{figure}[htbp]
  \centering
  \includegraphics[width=\linewidth]{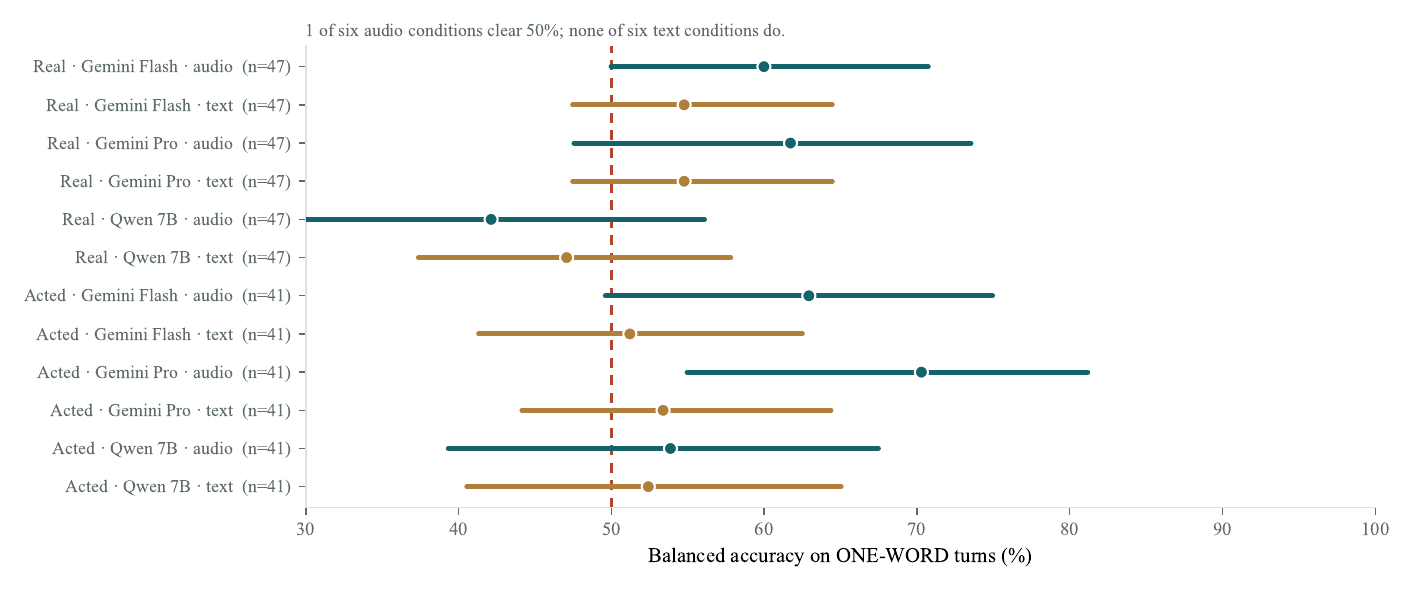}
  \caption{One-word turns: not one transcript-only condition beats a coin flip. Every model and input, with \SI{95}{\percent} Wilson intervals against the chance line. Six transcript conditions, none excluding \SI{50}{\percent}.}
  \label{fig:oneword}
\end{figure}

\begin{figure}[htbp]
  \centering
  \includegraphics[width=\linewidth]{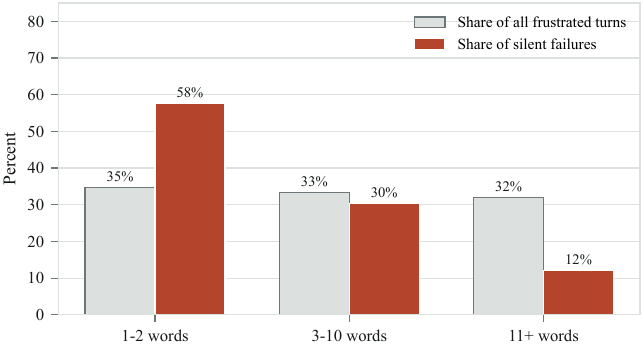}
  \caption{Silent failures concentrate in short turns for the transcript condition. Turns wrongly cleared, by turn length. The transcript condition's failures concentrate in the shortest turns.}
  \label{fig:failures}
\end{figure}

\subsection{Accuracy and intervals by turn length}
\label{app:bylength}

Accuracy split by turn length for every model and input, with intervals and the class counts in each bucket.

Table~\ref{tab:oneword} gives recall on one-word turns, the bucket the short-turn
claims rest on. Recall is the share of frustrated turns a condition actually
flagged, so unlike balanced accuracy it says whether a system would have raised
anything at all.

Every transcript condition on the clinical calls recovers two of the \num{21}
frustrated one-word turns, and it is the same two in all three cases. Audio on
the same turns ranges from \SI{23.8}{\percent} to \SI{61.9}{\percent}, better by
a wide margin and still short of what a monitor needs. The acted corpus behaves
the same way, with one exception: the open model's transcript condition reaches
\SI{22.2}{\percent} there, roughly double either Gemini model, which is the only
one-word transcript cell in the study that rises meaningfully above the floor.

\begin{table}[htbp]
  \caption{Every condition on one-word turns, the bucket the short-turn claims
    rest on. Balanced accuracy with its \SI{95}{\percent} Wilson interval, and
    recall as the share of frustrated one-word turns the condition flagged
    (\num{21} such turns on the clinical set, \num{18} on the acted one). A tick
    marks the one condition whose interval excludes chance. The audio range
    quoted in Section~\ref{sec:c2} spans both models and corpora: its floor is
    the open model on clinical audio, which sits below chance, and its ceiling
    is Gemini 2.5 Pro on acted audio.}
  \label{tab:oneword}
  \centering
  \small
  \begin{tabular}{lllcccc}
    \toprule
    Corpus & Model & Input & Bal.\ acc.\ & \SI{95}{\percent} CI & Recall (count) & Excludes chance \\
    \midrule
    Clinical & Gemini 2.5 Flash & Text & 54.8 & [47.5, 64.5] & 9.5\,(2/21) &  \\
    Clinical & Gemini 2.5 Pro & Text & 54.8 & [47.5, 64.5] & 9.5\,(2/21) &  \\
    Clinical & Qwen2.5-Omni 7B & Text & 47.1 & [37.4, 57.8] & 9.5\,(2/21) &  \\
    Acted & Gemini 2.5 Flash & Text & 51.2 & [41.3, 62.5] & 11.1\,(2/18) &  \\
    Acted & Gemini 2.5 Pro & Text & 53.4 & [44.1, 64.4] & 11.1\,(2/18) &  \\
    Acted & Qwen2.5-Omni 7B & Text & 52.4 & [40.5, 65.0] & 22.2\,(4/18) &  \\
    \addlinespace[3pt]
    Clinical & Gemini 2.5 Flash & Audio & 60.0 & [50.0, 70.7] & 23.8\,(5/21) &  \\
    Clinical & Gemini 2.5 Pro & Audio & 61.7 & [47.6, 73.5] & 61.9\,(13/21) &  \\
    Clinical & Qwen2.5-Omni 7B & Audio & 42.1 & [29.8, 56.1] & 38.1\,(8/21) &  \\
    Acted & Gemini 2.5 Flash & Audio & 62.9 & [49.6, 74.9] & 38.9\,(7/18) &  \\
    Acted & Gemini 2.5 Pro & Audio & 70.3 & [55.0, 81.2] & 66.7\,(12/18) & \checkmark \\
    Acted & Qwen2.5-Omni 7B & Audio & 53.9 & [39.3, 67.5] & 55.6\,(10/18) &  \\
    \bottomrule
  \end{tabular}
\end{table}

\begin{table}[htbp]
  \caption{Turn counts by utterance length for both evaluation sets, with the
    frustrated and not-frustrated split in each bucket.}
  \label{tab:lengthn}
  \centering
  \small
  \begin{tabular}{lcccc}
    \toprule
    & \multicolumn{2}{c}{Clinical} & \multicolumn{2}{c}{Acted} \\
    \cmidrule(lr){2-3}\cmidrule(lr){4-5}
    Turn length & $n$ & frustrated / not & $n$ & frustrated / not \\
    \midrule
    1 word  &  47 & 21 / 26 &  41 & 18 / 23 \\
    2--3    &  24 &  8 / 16 &  20 &  4 / 16 \\
    4--5    &  10 &  5 / 5  &  11 &  5 / 6  \\
    6--10   &  34 & 17 / 17 &  28 & 14 / 14 \\
    11--20  &  22 & 16 / 6  &  24 & 14 / 10 \\
    21+     &  13 &  8 / 5  &  26 & 21 / 5  \\
    \midrule
    All     & 150 & 75 / 75 & 150 & 76 / 74 \\
    \bottomrule
  \end{tabular}
\end{table}

Table~\ref{tab:spread} gives the quantity Section~\ref{sec:shortcut} quotes:
how far each condition's predicted-frustration rate travels across the six
length buckets, against how far the true rate itself travels. Figure~\ref{fig:lengthbias}
plots the same curves. A condition tracking frustration rather than length
would move about as much as the truth does. On the clinical calls every
transcript condition moves at least \SI{11}{pp} further than the truth, and the
audio conditions for Gemini 2.5 Pro and the open model move less than it.

\begin{table}[htbp]
  \caption{Spread of the predicted-frustration rate across the six turn-length
    buckets, in percentage points, against the spread of the true rate. Larger
    than the true spread means a condition varies with length more than
    frustration itself does.}
  \label{tab:spread}
  \centering
  \small
  \begin{tabular}{llcc}
    \toprule
    Input & Model & Clinical & Acted \\
    \midrule
    \multicolumn{2}{l}{\itshape True rate} & 39.4 & 60.8 \\
    \midrule
    Text  & Gemini 2.5 Flash & 72.7 & 74.9 \\
    Text  & Gemini 2.5 Pro   & 88.1 & 77.3 \\
    Text  & Qwen2.5-Omni 7B  & 50.9 & 70.8 \\
    \addlinespace[2pt]
    Audio & Gemini 2.5 Flash & 66.9 & 78.5 \\
    Audio & Gemini 2.5 Pro   & 32.9 & 58.5 \\
    Audio & Qwen2.5-Omni 7B  & 20.8 & 55.8 \\
    \bottomrule
  \end{tabular}
\end{table}

\begin{figure}[htbp]
  \centering
  \includegraphics[width=\linewidth]{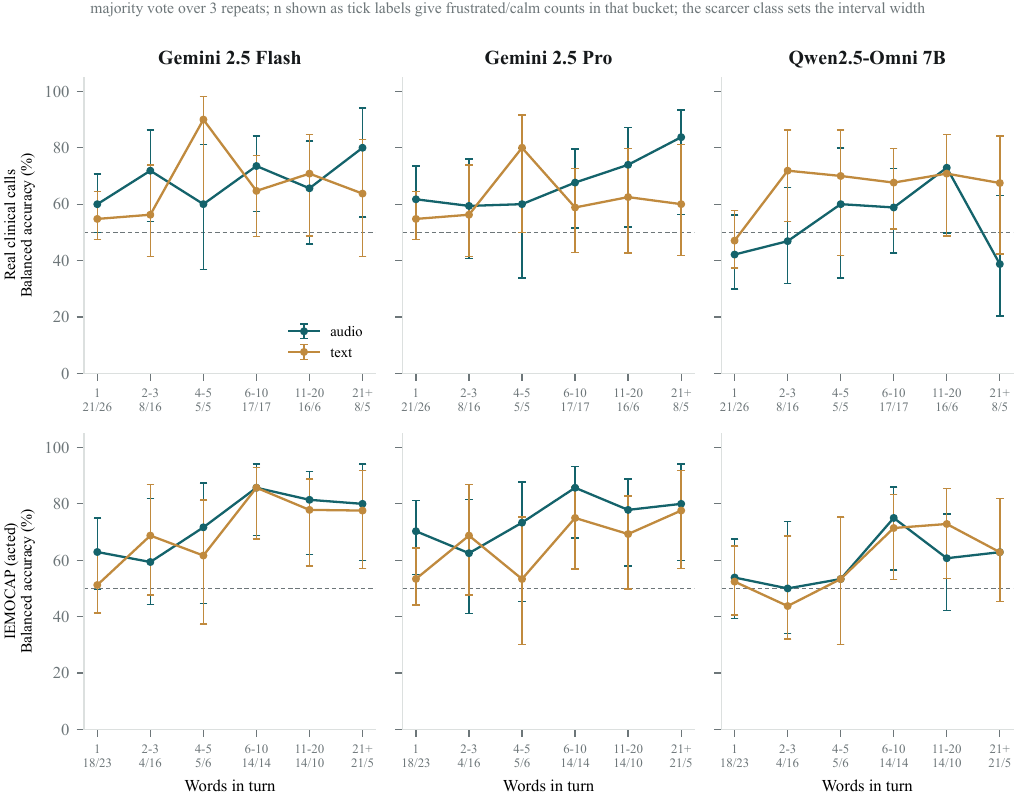}
  \caption{Balanced accuracy by turn length, with \SI{95}{\percent} Wilson intervals, for every model and input. Tick labels give the frustrated and not-frustrated counts in each bucket: the scarcer of the two sets the interval width, which is why some larger buckets are less precisely measured than smaller ones.}
  \label{fig:cilength}
\end{figure}
\subsection{Direction of text error by utterance length}
\label{app:lenerr}

Splitting the transcript condition's errors by direction separates two failures
that a single accuracy figure combines. A system that judged a short turn and got
it wrong would produce misses and false alarms in some proportion. This one
produces almost only misses: on one-word turns neither Gemini model raises a false
alarm, and the open model raises \num{4}, which is the signature of a condition that is declining to commit rather than
committing incorrectly.

\begin{figure}[htbp]
  \centering
  \includegraphics[width=0.78\linewidth]{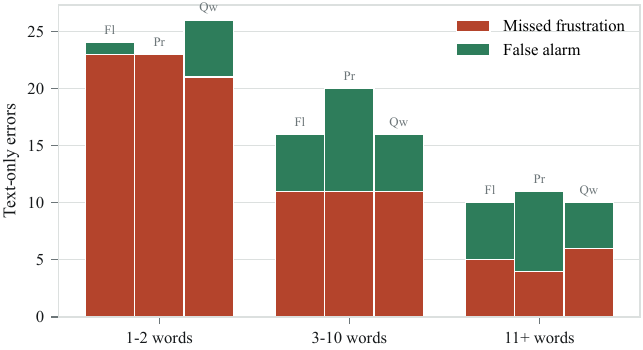}
  \caption{Text errors are almost all misses on short turns, split into missed frustration and false alarms. On
    one-word clinical turns the three transcript conditions produce \num{57} misses
    (\num{19} each) and \num{4}
    false alarms, all from the open model: it is not judging those turns and getting them wrong, it is
    declining to judge them, and in a binary decision declining reads as calm.}
  \label{fig:lenerr}
\end{figure}

\subsection{Direction of error on short turns}
\label{app:errfig}

The same asymmetry holds across most models and inputs. On the \num{55} short
turns, the three transcript conditions and Gemini 2.5 Flash on audio err almost
entirely by missing frustration, with \numrange{18}{23} misses against
\numrange{0}{5} false alarms. Gemini 2.5 Pro and the open model on audio err in
both directions about equally: \num{11} misses and \num{11} false alarms, and
\num{17} and \num{15}. A monitor built on the first behaviour clears turns it
should escalate; one built on the second escalates almost as often as it misses.

\begin{figure}[htbp]
  \centering
  \includegraphics[width=0.84\linewidth]{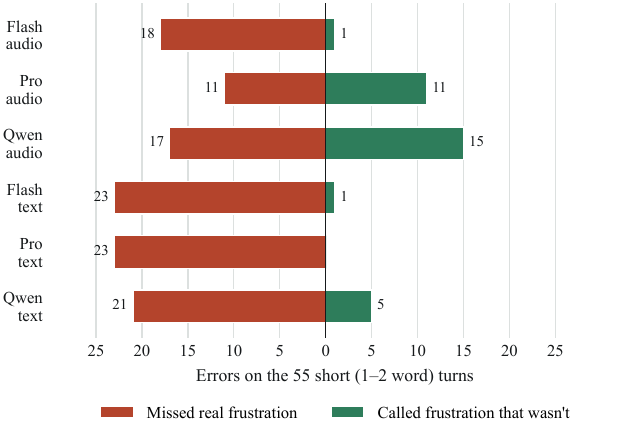}
  \caption{Direction of error on the \num{55} short turns. Four of six
    model--modality conditions err almost exclusively by \emph{missing} real
    frustration. Gemini 2.5 Pro and the open model on audio err in both
    directions about equally. Neither failure mode is usable for monitoring.}
  \label{fig:errors}
\end{figure}

\subsection{Clearing behaviour by model, threshold and call}
\label{app:clearing}

A turn's frustration score, for a given input, is the share of votes calling it
frustrated, pooled over the two frontier models and their three repeats each:
six votes per turn per input. The open model takes no part in it. A turn is
\emph{cleared at threshold} $\tau$ when that share is at most $\tau$, and the
\emph{confidently cleared} counts in Section~\ref{sec:safety} are the case
$\tau=0$, where not one of the six votes called the turn frustrated.

Two objections to the clearing result are that it depends on where that
threshold is drawn, and that it is carried by one unusual call. Neither
survives. We sweep $\tau$ over $\{0,\,0.1,\,0.2,\,0.25,\,0.34,\,0.5\}$, which
with six votes resolves to four distinct cut points, and hold out each of the
six calls in turn. Loosening the threshold makes the transcript condition clear
more turns without making the turns it clears any safer: it goes from clearing
\num{87} turns at $\tau=0$ to \num{104} at $\tau=0.5$ while the share of those
that were in fact frustrated stays between \SI{37.5}{\percent} and
\SI{40.0}{\percent} throughout. Audio clears fewer turns at every threshold and
is wrong about fewer of them, rising from \SI{17.8}{\percent} to
\SI{31.7}{\percent} as the criterion weakens. Holding out each call in turn
leaves both conditions close to their full-sample values.

\begin{figure}[htbp]
  \centering
  \includegraphics[width=\linewidth]{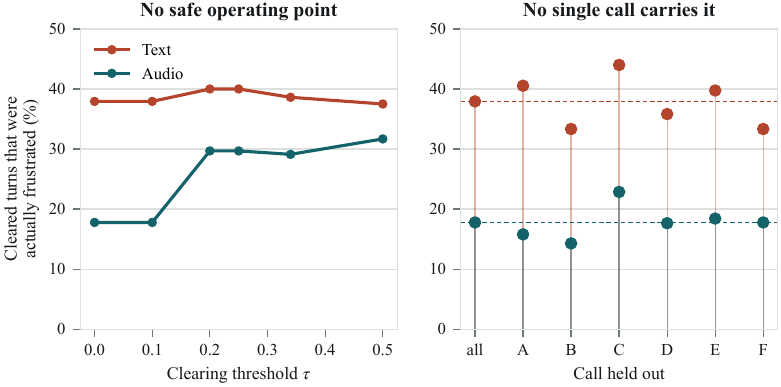}
  \caption{The clearing gap is neither a threshold artefact nor the product of
    one call. \emph{Left}: sweeping the clearing threshold $\tau$ leaves text
    between \SI{37.5}{\percent} and \SI{40.0}{\percent} wrong, so loosening
    $\tau$ makes text clear more turns without making it safer; audio rises from
    \SI{17.8}{\percent} to \SI{31.7}{\percent}, closing much of the gap as the
    criterion weakens. \emph{Right}: holding out each call in turn moves neither
    condition far from its full-sample value, shown dashed.}
  \label{fig:robust}
\end{figure}

\section{Supporting the input comparison}

\begin{figure}[htbp]
  \centering
  \includegraphics[width=\linewidth]{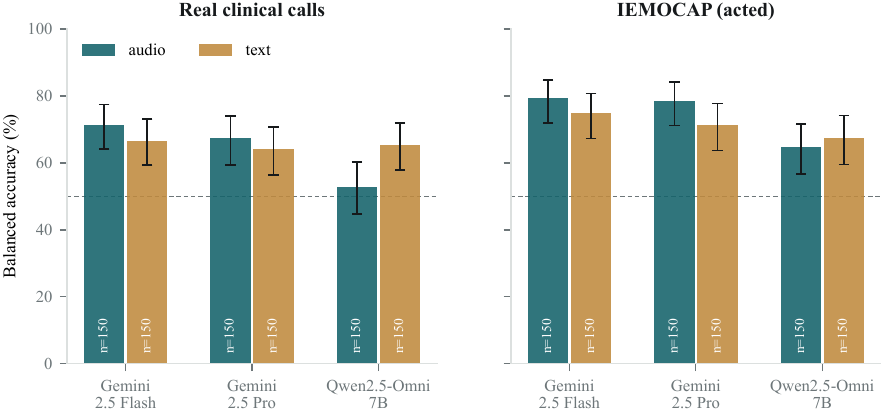}
  \caption{Balanced accuracy for every model and input on both corpora, with \SI{95}{\percent} Wilson intervals and the number of turns on each bar. Interval widths differ because a proportion's variance peaks at one half, so cells nearer chance are measured less precisely.}
  \label{fig:headline}
\end{figure}

The pooled difference between the two inputs is a summary of the turns on which
they disagree, and both the size and the direction of that set vary by model. On
the clinical calls audio answers \num{17} turns correctly that the transcript
misses for Gemini 2.5 Flash and \num{25} for Gemini 2.5 Pro, against \num{10} and
\num{20} in the other direction; for the open model the counts reverse, \num{19}
against \num{38}.

\begin{figure}[htbp]
  \centering
  \includegraphics[width=\linewidth]{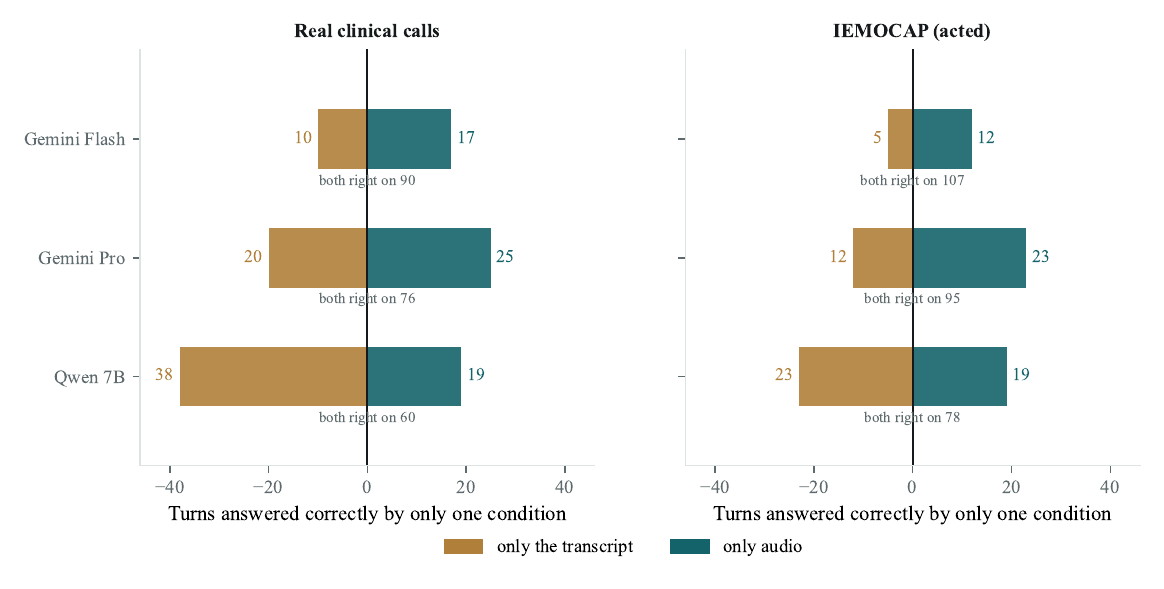}
  \caption{The paired evidence: turns exactly one input answered correctly, with the count both answered correctly beneath. The two inputs score identical turns, so this is the comparison the pooled difference summarises.}
  \label{fig:discordance}
\end{figure}

\subsection{Complementarity and fusion}
\label{app:compl}

\begin{figure}[htbp]
  \centering
  \includegraphics[width=\linewidth]{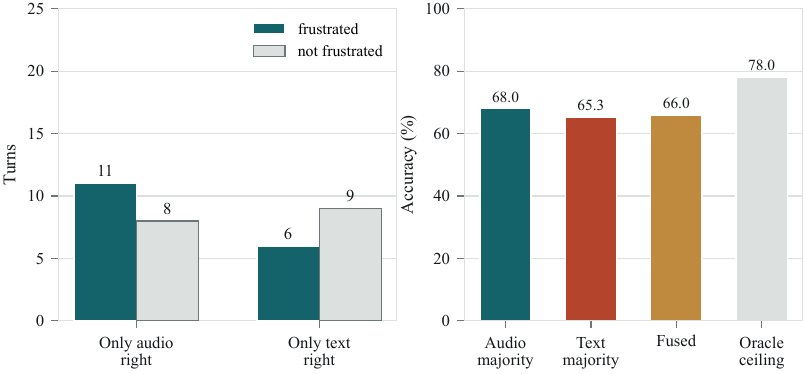}
  \caption{\emph{Left}: of the \num{19} turns only audio gets right, \num{11}
    are frustrated; of the \num{15} turns only text gets right, \num{9} are
    calm. \emph{Right}: a simple vote across the models and inputs reaches
    \SI{66.0}{\percent}, against an oracle ceiling of \SI{78.0}{\percent}
    from picking the better input for each turn.}
  \label{fig:compl}
\end{figure}

Because the two inputs fail on different turns rather than on the same ones, it is
worth asking what combining them could achieve. A simple vote across the models
and inputs reaches \SI{66.0}{\percent}, below either frontier model's audio
alone, while an oracle that picked the better input for each turn individually
bounds what any combination rule could reach (Figure~\ref{fig:compl}).

\subsection{The agent's preceding turn as context}
\label{app:agentturn}

We reran all three models with the agent's preceding turn, what the agent said
immediately before the patient spoke, added to the prompt in the same words for
both inputs; the prompt states that it is there only to show what the patient is
answering (Appendix~\ref{app:prompts}). The agent's turn is the agent audio between the end
of the previous patient turn and the start of this one, capped at the last
\SI{30}{\second}. Where that gap is silent, because the patient paused and went
on, the agent's most recent voiced turn is used instead. One turn has no agent
speech before it, and its prompt carries the line
\texttt{[no agent speech captured]} in its place. The
agent's turns were transcribed with Whisper large-v3 and every one was checked
against the audio by ear. Each model ran three repeats, scored by majority as
in the main runs. No transcript condition changes by an interval that
excludes zero, and one-word transcript recall stays between \num{3} and \num{5}
of \num{21} (Table~\ref{tab:agentturn}). Outside the call that holds most
frustrated one-word turns the agent's turn helps Gemini 2.5 Pro's transcript more,
from \num{0} to \num{4} of \num{9}, on too few turns to estimate. With the
agent's turn, the frontier pair's transcript clears \num{83} turns of which
\SI{32.5}{\percent} were frustrated, against \num{52} and \SI{19.2}{\percent}
for audio. Within the run, audio minus transcript is \SI{0.0}{pp} for Gemini 2.5
Flash, \SI{6.7}{pp} for Gemini 2.5 Pro and \SI{-6.0}{pp} for Qwen2.5-Omni, against
\SI{4.7}{pp}, \SI{3.3}{pp} and \SI{-12.7}{pp} on the patient turn only; only the last
of these six has an interval excluding zero (Table~\ref{tab:agentturndiff}).

\begin{table}[htbp]
  \caption{The agent's preceding turn added as context, on the clinical
    set, beside the main results on the patient turn only. Balanced accuracy with its
    \SI{95}{\percent} Wilson interval; the paired change from the main results,
    with its paired-bootstrap interval; frustrated and not-frustrated turns
    classified correctly, of \num{75} each; and recall on the \num{21}
    frustrated one-word turns, overall and on the \num{9} outside the call that
    holds the other \num{12}.}
  \label{tab:agentturn}
  \centering
  \small
  \resizebox{\linewidth}{!}{%
  \begin{tabular}{lllcccccc}
    \toprule
    Run & Model & Input & Bal.\ acc.\ [95\% CI] & Change [95\% CI] & Frustrated & Not frustrated & One-word & One-word, other calls \\
    \midrule
    Patient turn only & Gemini 2.5 Flash & Text  & 66.7 [59.4, 73.1] & -- & 36/75 & 64/75 & 2/21 & 0/9 \\
                     & Gemini 2.5 Pro   & Text  & 64.0 [56.4, 70.8] & -- & 37/75 & 59/75 & 2/21 & 0/9 \\
                     & Qwen2.5-Omni 7B  & Text  & 65.3 [57.8, 71.9] & -- & 37/75 & 61/75 & 2/21 & 0/9 \\
                     & Gemini 2.5 Flash & Audio & 71.3 [64.1, 77.4] & -- & 41/75 & 66/75 & 5/21 & 1/9 \\
                     & Gemini 2.5 Pro   & Audio & 67.3 [59.4, 74.1] & -- & 49/75 & 52/75 & 13/21 & 2/9 \\
                     & Qwen2.5-Omni 7B  & Audio & 52.7 [44.8, 60.3] & -- & 36/75 & 43/75 & 8/21 & 2/9 \\
    \midrule
    With agent's preceding turn & Gemini 2.5 Flash & Text  & 65.3 [58.3, 71.6] & $-1.3$ [$-6.7$, $+4.0$]  & 32/75 & 66/75 & 3/21 & 2/9 \\
                     & Gemini 2.5 Pro   & Text  & 65.3 [57.6, 72.1] & $+1.3$ [$-5.3$, $+8.0$]  & 40/75 & 58/75 & 5/21 & 4/9 \\
                     & Qwen2.5-Omni 7B  & Text  & 63.3 [55.7, 70.2] & $-2.0$ [$-9.3$, $+5.3$]  & 37/75 & 58/75 & 3/21 & 1/9 \\
                     & Gemini 2.5 Flash & Audio & 65.3 [58.4, 71.5] & $-6.0$ [$-12.7$, $+0.7$] & 31/75 & 67/75 & 3/21 & 2/9 \\
                     & Gemini 2.5 Pro   & Audio & 72.0 [64.4, 78.2] & $+4.7$ [$-3.3$, $+12.7$] & 47/75 & 61/75 & 15/21 & 5/9 \\
                     & Qwen2.5-Omni 7B  & Audio & 57.3 [49.7, 64.5] & $+4.7$ [$-5.3$, $+14.7$] & 32/75 & 54/75 & 1/21 & 1/9 \\
    \bottomrule
  \end{tabular}}
\end{table}

\begin{table}[htbp]
  \caption{Audio minus transcript in balanced accuracy (pp) on the clinical set,
    on the patient turn only and with the agent's preceding turn, with paired-bootstrap
    \SI{95}{\percent} intervals.}
  \label{tab:agentturndiff}
  \centering
  \small
  \begin{tabular}{lcc}
    \toprule
    Model & Patient turn only & With agent's preceding turn \\
    \midrule
    Gemini 2.5 Flash & $+4.7$ [$-2.0$, $+11.3$]  & $+0.0$ [$-6.0$, $+6.0$] \\
    Gemini 2.5 Pro   & $+3.3$ [$-5.3$, $+12.0$]  & $+6.7$ [$-1.3$, $+14.7$] \\
    Qwen2.5-Omni 7B  & $-12.7$ [$-22.0$, $-3.3$] & $-6.0$ [$-14.0$, $+2.0$] \\
    \bottomrule
  \end{tabular}
\end{table}

\subsection{Context depth pooled across utterance lengths}
\label{app:pooled}

Swept over the whole evaluation set, context depth has no visible effect.
Pooled across lengths the curve is flat and no depth is distinguishable from
another: the variation across repeats at a fixed depth is comparable to the
variation across depths. Stratified by length (Section~\ref{sec:whytext},
Appendix~\ref{app:contextfig}), the two length strata move in opposite
directions and cancel when pooled. We report it because it illustrates a
general hazard: an aggregate context-depth ablation can conceal a large effect of
opposite sign in two subpopulations.

\subsection{Context depth by utterance length}
\label{app:contextfig}

\begin{figure}[htbp]
  \centering
  \includegraphics[width=\linewidth]{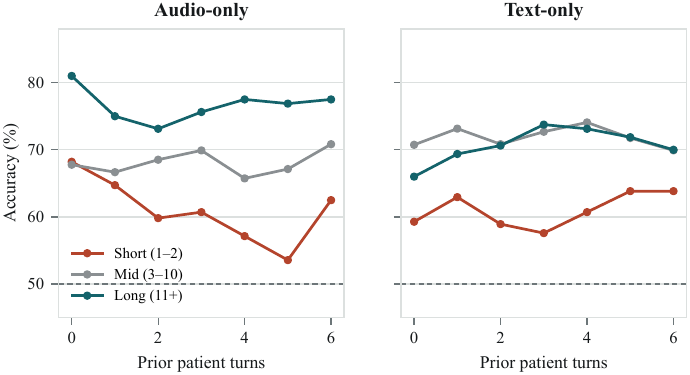}
  \caption{Supplying preceding patient audio degrades short-turn accuracy while
    leaving long turns intact; text moves in the opposite direction and gains
    little.}
  \label{fig:context}
\end{figure}

\subsection{Model capability and the input gap}
\label{app:modelcap}

The input comparison as differences with intervals, so the paired gap and its precision are visible together.

\begin{figure}[htbp]
  \centering
  \includegraphics[width=\linewidth]{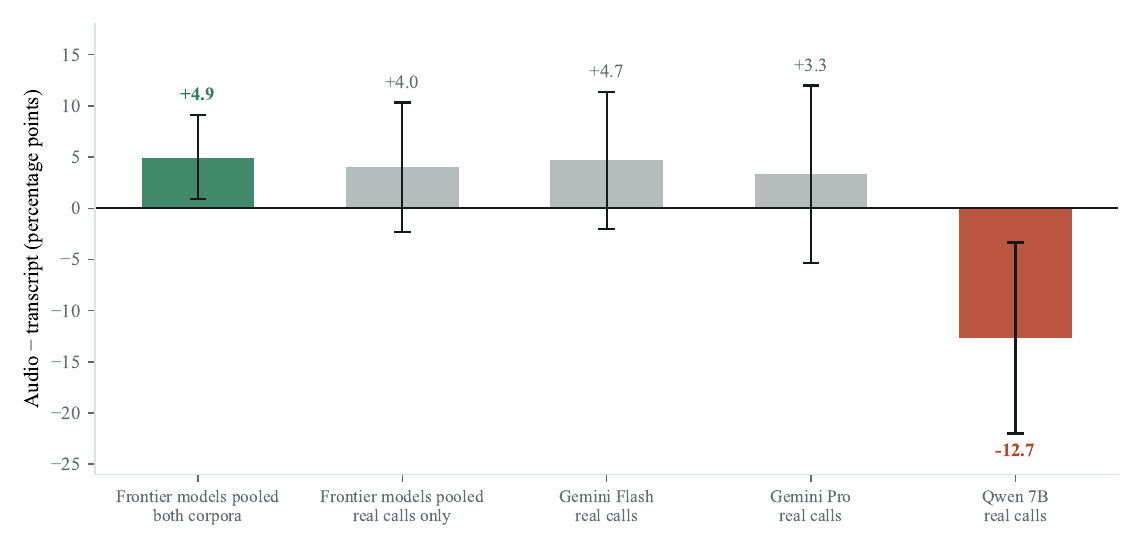}
  \caption{Does hearing the voice help? The modality comparisons that matter, as differences with \SI{95}{\percent} intervals. Solid bars exclude zero.}
  \label{fig:moddiff}
\end{figure}

\section{Speech-encoder baselines}
\label{app:probes}

\subsection{A trained acoustic baseline}
\label{app:probetransfer}

This appendix gives the full results for the probes trained on acted speech,
which Section~\ref{sec:probes} summarises, and the training details for the
probes trained on the 30 calls. It first asks what the corpus gap looks like for
a classifier fitted on speech.

Frozen WavLM Base+ representations are taken from a single intermediate layer
(one of layers 6--9 of 12) and pooled over the turn by mean, maximum, or mean and
standard deviation, with either a logistic regression or an RBF
support vector machine fitted on top. Layer, pooling and hyperparameters are
chosen by cross-validation inside the training data alone, and nothing in WavLM
is fine-tuned. Each configuration is fitted once on \num{1810} IEMOCAP turns,
split from its test set by dialogue so that no dialogue contributes to both, and
is then scored twice without refitting: on \num{150} held-out acted turns and on
the \num{150} real clinical turns, which take no part in training or in
selection. Two bandwidth arms are reported, since the corpora differ in channel
as well as in content. The native arm trains on the acted audio as recorded at
\SI{16}{\kilo\hertz}; the band-matched arm trains on the same turns after an
\SI{8}{\kilo\hertz} telephone round trip, so that training and test share an
effective bandwidth.

Table~\ref{tab:appendix_generalisation} reports the result. A probe given enough
acted speech learns something the transcript does not: the strongest
configuration, WavLM with an SVM on native audio, reaches \SI{70.0}{\percent} on
the clinical turns against \SI{58.0}{\percent} for word count and
\SI{52.7}{\percent} for clip duration. What does not go away is the corpus gap.
Every configuration scores lower on the clinical turns than on held-out acted
speech, by between \SI{6.1}{pp} and \SI{11.3}{pp}, and the band-matched arms drop
as far as the native ones, so the loss is not carried by bandwidth. Word count
transfers more evenly, falling only \SI{4.3}{pp}, which is what a predictor with no access to the speech should do; clip duration falls \SI{8.2}{pp}.

\subsection{Separating training from test}
\label{app:probesplit}

Splitting training from test by dialogue is the arrangement reported throughout. Splitting by speaker instead trains on more turns but leaves only \num{42} acted turns to test on, so its acted column is measured far less precisely.

\begin{figure}[htbp]
  \centering
  \includegraphics[width=\linewidth]{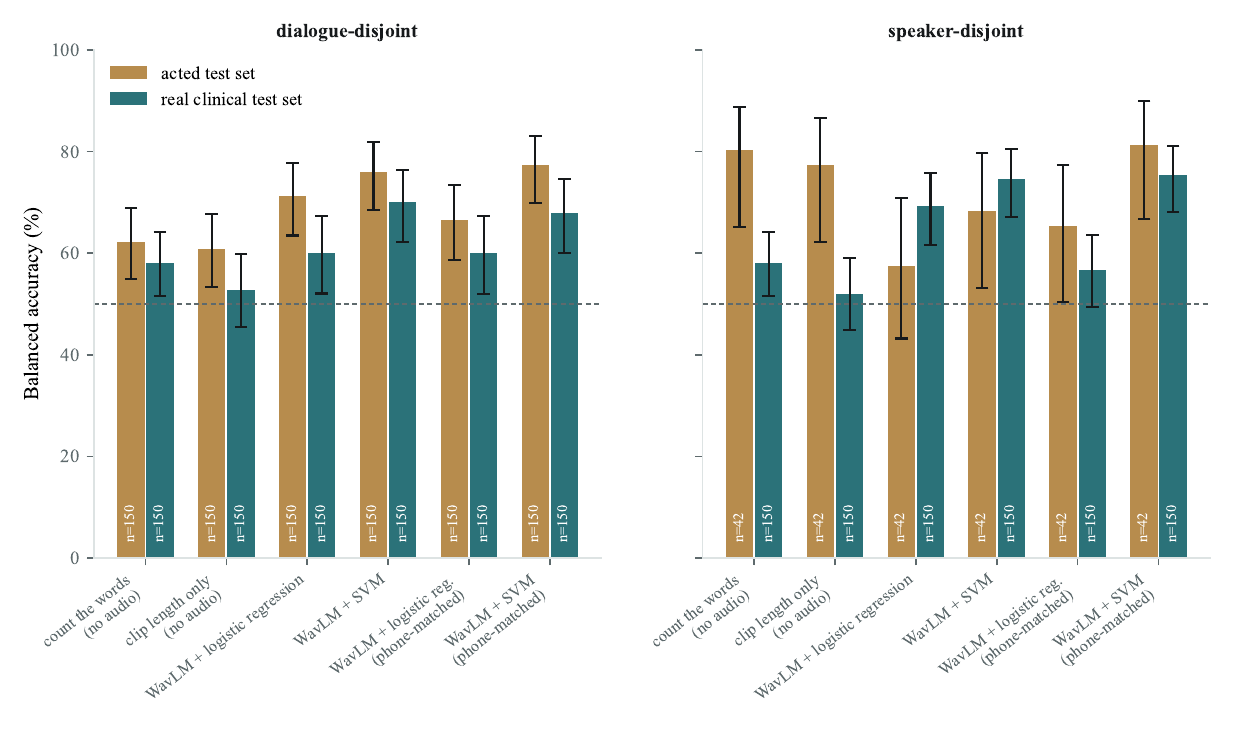}
  \caption{The probes and the audio-free baselines on both test sets, with \SI{95}{\percent} intervals and the number of turns on each bar. \emph{Left}: training and test split by dialogue, the arrangement Table~\ref{tab:appendix_generalisation} reports. \emph{Right}: split by speaker, which trains on more turns but tests the acted side on only \num{42}.}
  \label{fig:poolregimes}
\end{figure}

\subsection{Training-set size}
\label{app:probesize}

The probes' margin over counting words depends on how much labelled acted speech they are given.

\begin{figure}[htbp]
  \centering
  \includegraphics[width=\linewidth]{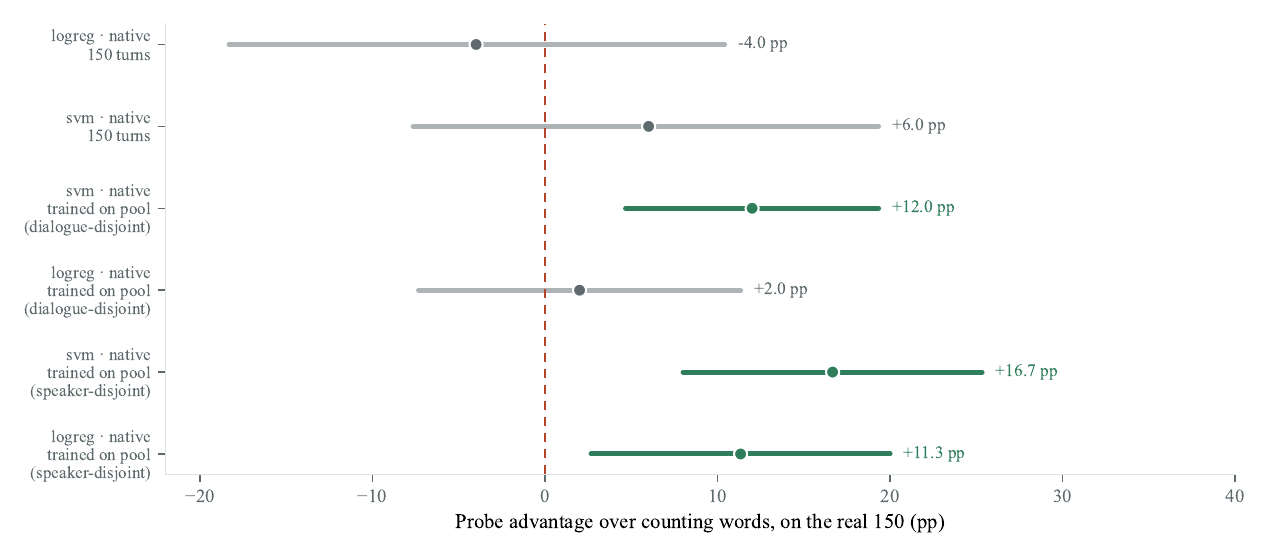}
  \caption{More training data turns the probes from no better than word count into clearly better. Probe advantage over counting words, trained on the labelled \num{150} against the full acted pool. The two \num{150}-turn rows use a deliberately conservative interval: that run did not save per-item predictions, so its probe and baseline cannot be compared on matched turns. The pool rows are properly paired.}
  \label{fig:probesize}
\end{figure}

\begin{table*}[t]
\centering
\caption{Generalisation from acted IEMOCAP speech to real clinical speech.
All values are balanced accuracy (\%), so chance is 50\%. Models were trained on the larger IEMOCAP pool.
Dialogue-disjoint evaluation uses a single held-out acted set and clinical set
($n=150$ each; $n_{\mathrm{train}}=1{,}810$). Session-rotation evaluation pools
predictions from five held-out-session folds for the acted set and reports the
mean and standard deviation across the five models for the same clinical set.}
\label{tab:appendix_generalisation}
\small
\begin{tabular}{llccc}
\toprule
\textbf{Features} & \textbf{Classifier}
& \multicolumn{2}{c}{\textbf{Dialogue-disjoint}}
& \textbf{Session rotation} \\
\cmidrule(lr){3-4}\cmidrule(l){5-5}
& & \textbf{Acted} & \textbf{Clinical}
& \textbf{Acted / Clinical} \\
\midrule
WavLM, band-matched & SVM
& 77.4 & 68.0 & 78.0 / 74.0 {\scriptsize$\pm$ 0.9} \\
WavLM, native & SVM
& 76.1 & 70.0 & 73.4 / 71.3 {\scriptsize$\pm$ 2.7} \\
WavLM, native & Logistic regression
& 71.3 & 60.0 & 66.6 / 64.8 {\scriptsize$\pm$ 4.0} \\
WavLM, band-matched & Logistic regression
& 66.6 & 60.0 & 64.0 / 62.7 {\scriptsize$\pm$ 4.0} \\
\midrule
Word count & Logistic regression
& 62.3 & 58.0 & 62.3 / 57.9 {\scriptsize$\pm$ 0.3} \\
Duration & Logistic regression
& 60.9 & 52.7 & 60.2 / 52.3 {\scriptsize$\pm$ 0.4} \\
\bottomrule
\end{tabular}
\end{table*}

\begin{table}[t]
\centering
\caption{Effect of acted training-set size on balanced accuracy for the same
held-out clinical test set ($n=150$). The smaller-data condition uses 150 acted training
turns; the pooled condition uses 1,810 dialogue-disjoint acted turns.
$\Delta$ denotes the change in percentage points.}
\label{tab:appendix_training_size}
\small
\begin{tabular}{llrrr}
\toprule
\textbf{Features} & \textbf{Classifier}
& \textbf{$n=150$} & \textbf{$n=1{,}810$} & \textbf{$\Delta$} \\
\midrule
WavLM, band-matched & SVM
& 56.7 & 68.0 & +11.3 \\
WavLM, native & SVM
& 66.7 & 70.0 & +3.3 \\
WavLM, native & Logistic regression
& 56.7 & 60.0 & +3.3 \\
WavLM, band-matched & Logistic regression
& 61.3 & 60.0 & -1.3 \\
\midrule
Word count & Logistic regression
& 60.7 & 58.0 & -2.7 \\
\bottomrule
\end{tabular}
\end{table}

Increasing the acted training set from 150 to 1,810 turns improved clinical accuracy by up to 11.3 percentage points for the WavLM probes, whereas the word-count baseline remained broadly unchanged. This indicates that the relatively small advantage of the probes in the earlier experiment was partly attributable to limited training data. Under five-fold held-out-session rotation, the strongest configuration achieved 74.0\% clinical accuracy, compared with \SI{57.9}{\percent} for the word-count baseline, and exhibited an acted-to-clinical gap of 4.0 percentage points.

\subsection{One-word turns and the decision threshold}
\label{app:probethreshold}

The probes set their decision threshold on acted speech. To check that this
cut-off does not flatter them on the clinical calls, we also chose it on
clinical speech while leaving one call out: each call is scored with the
threshold that maximises balanced accuracy on the other five. For the strongest
configuration, band-matched WavLM with an SVM under session rotation, balanced
accuracy averages \SI{74.0}{\percent} and the probe flags \num{14.0} of the
\num{21} frustrated one-word turns on average under either threshold, though
both vary from fold to fold (Table~\ref{tab:probethreshold}).
Section~\ref{sec:probes} reports this probe on the 24 added calls.

\begin{table}[htbp]
  \caption{WavLM SVM probes on the \num{150} clinical turns with the
    threshold set on acted speech or on the other clinical calls (leave one
    call out). Balanced accuracy and the frustrated one-word turns flagged, of
    \num{21}. Band-matched features use the five session-rotation folds; the
    native-band row is a single model.}
  \label{tab:probethreshold}
  \centering
  \small
  \begin{tabular}{lcccc}
    \toprule
    & \multicolumn{2}{c}{Acted threshold} & \multicolumn{2}{c}{Held-out-call threshold} \\
    \cmidrule(lr){2-3}\cmidrule(lr){4-5}
    & Bal.\ acc. & One-word & Bal.\ acc. & One-word \\
    \midrule
    Band-matched, fold 1 & 74.0 & 11/21 & 72.0 & 9/21 \\
    Band-matched, fold 2 & 72.7 & 10/21 & 71.3 & 9/21 \\
    Band-matched, fold 3 & 74.0 & 14/21 & 72.0 & 15/21 \\
    Band-matched, fold 4 & 74.0 & 16/21 & 76.0 & 18/21 \\
    Band-matched, fold 5 & 75.3 & 19/21 & 78.7 & 19/21 \\
    Band-matched, mean   & 74.0 & 14.0/21 & 74.0 & 14.0/21 \\
    \midrule
    Native band          & 70.0 & 4/21 & 66.0 & 9/21 \\
    \bottomrule
  \end{tabular}
\end{table}

\subsection{Probes trained on the 30 calls}
\label{app:probethirty}

The probes of Section~\ref{sec:probes} that are trained on clinical speech use the same recipe as those trained on acted speech: frozen WavLM layers pooled over the turn by mean, maximum, or mean and standard deviation, with a logistic regression or an RBF support vector machine. No call outside the 30 is used for training or for choosing any setting. The 30 calls are split into five folds by patient, and each fold's turns are scored by a probe whose layer, pooling and weights come from the other four folds only. Layer and pooling are chosen by cross-validation grouped by patient inside those four folds. A turn is flagged at the probe's own decision boundary. Three split seeds give three predictions per turn, and the turn's prediction is their majority, as for the language models. Selection settles on WavLM layers 6 to 9 for both probes, mostly with mean pooling for the SVM and with all three poolings for the logistic regression. The three seeds disagree on \SI{29.1}{\percent} of the \num{714} turns for the logistic regression and \SI{13.3}{\percent} for the SVM.

\section{Can these parameters be measured on a one-word turn?}
\label{app:measure}

Every functional in eGeMAPS summarises a window of speech, so its value carries sampling
noise that grows as the window shortens. We estimate that noise directly and without
labels. For each turn of at least six seconds we compute a parameter over the whole turn,
then recompute it on fifteen randomly placed windows drawn from inside the same turn. The
turn is unchanged, so variation across its own windows is measurement error; comparing it
against variation between turns gives a one-way random-effects intraclass correlation.

\begin{figure}[htbp]
  \centering
  \includegraphics[width=\linewidth]{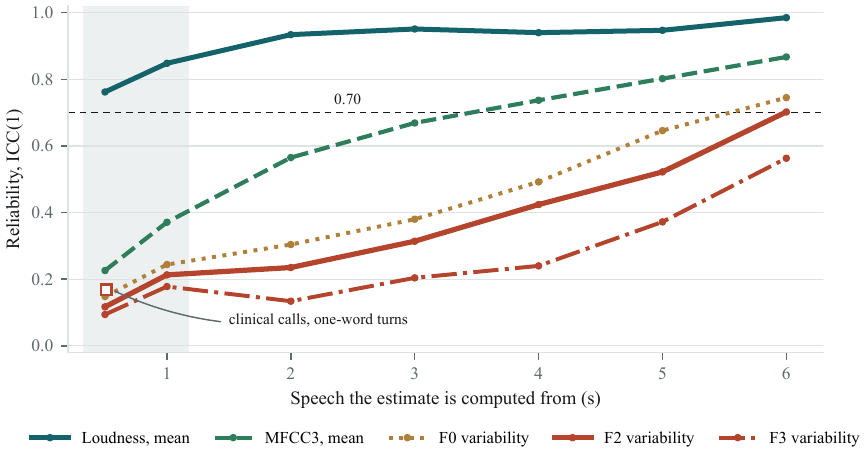}
  \caption{Reliability of eGeMAPS parameters against the amount of speech the estimate is
    computed from. Lines are the IEMOCAP window sweep; squares mark the clinical calls at
    the median one-word duration, where the two corpora agree. Level parameters are
    reliable from half a second, variability parameters are not.
    Shaded band marks the one-word range, \SIrange{0.32}{1.18}{\second}. One-way
    random-effects ICC(1); \num{45} source utterances, \num{15} windows per utterance at
    each length.}
  \label{fig:measure}
\end{figure}

Figure~\ref{fig:measure} reports this across window lengths. Level parameters are reliable
from half a second: loudness mean reaches \num{0.76} there. Variability parameters are not,
and F2 frequency variability reaches conventional reliability only near six seconds. On the
clinical calls at \SI{0.5}{\second}, F2 and F3 variability reach \num{0.17}, and none of
the \num{88} parameters reaches \num{0.70}.

Two alternative explanations were tested and do not hold. Bandwidth is not the cause:
IEMOCAP at its native \SI{16}{\kilo\hertz} gives the same poor reliability at
\SI{0.5}{\second}, \num{0.19} and \num{0.17} for F2 and F3, and band-limiting it to a
\SIrange{300}{3400}{\hertz} telephone channel moves reliability slightly upward rather than
down. Nor is it an artefact of acted speech being acoustically uniform: the
between-utterance spread is comparable across the two corpora.

We also attempted the direct labelled comparison, contrasting frustrated against calm
one-word turns within call, and separately across turns of the identical word. Neither is
informative: at \num{42} and \num{16} turns respectively they detect only effects above
\num{0.97} and \num{1.57} standard deviations.

\section{Onset of frustration}
As well as the detection of frustration from real clinical calls, the prediction of the onset of frustration was also investigated. 
\subsection{The same word, escalating mid-call}
\label{app:escalation}

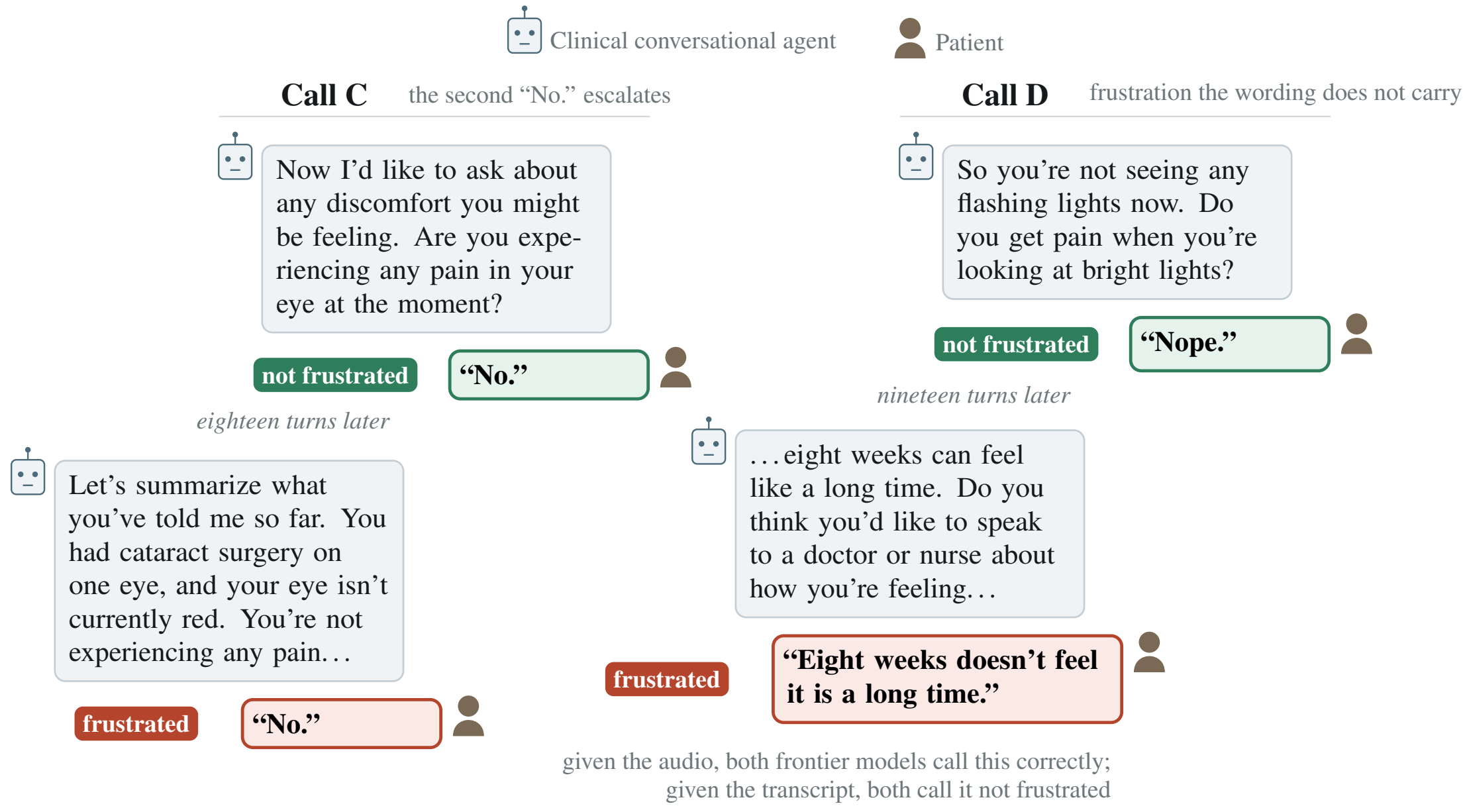
\begin{figure}[htbp]
  \centering
  \begin{tikzpicture}[
  font=\footnotesize,
  node distance=1.6mm,
  pics/patient/.style={code={
    \fill[patink] (0,0.10) circle (0.105);
    \fill[patink] (-0.155,-0.20) .. controls (-0.155,0.015) and (0.155,0.015)
                          .. (0.155,-0.20) -- cycle;
  }},
  pics/agent/.style={code={
    \draw[agentink,line width=0.45pt] (0,0.20) -- (0,0.285);
    \fill[agentink] (0,0.30) circle (0.028);
    \draw[agentink,line width=0.45pt,fill=agentbg,rounded corners=1.4pt]
      (-0.165,-0.145) rectangle (0.165,0.20);
    \fill[agentink] (-0.072,0.055) circle (0.028);
    \fill[agentink] ( 0.072,0.055) circle (0.028);
    \draw[agentink,line width=0.45pt] (-0.05,-0.055) -- (0.05,-0.055);
  }},
  bub/.style={rounded corners=3pt, inner sep=4pt, align=left,
              text width=32mm, line width=0.5pt},
  agentbub/.style={bub, draw=agentedge, fill=agentbg, text=ink, anchor=north west},
  patbub/.style  ={bub, draw=patedge,   fill=patbg,   text=ink, anchor=north east},
  hotbub/.style  ={bub, text width=17mm, draw=hotline,  fill=hotfill,  line width=0.9pt,
                   anchor=north east},
  coolbub/.style ={bub, text width=17mm, draw=coolline, fill=coolfill, line width=0.9pt,
                   anchor=north east},
  pill/.style={rounded corners=2.4pt, inner sep=2.4pt, font=\scriptsize\bfseries,
               text=white, anchor=east},
  head/.style={font=\small\bfseries, text=ink},
  verdict/.style={font=\scriptsize, text=labelgrey, anchor=north east, align=right},
  sub/.style={font=\scriptsize, text=labelgrey},
  leg/.style={font=\scriptsize, text=labelgrey, anchor=west, inner sep=0pt},
]

\pic at (3.05,0.66) {agent};
\node[leg] at (3.30,0.62) {Clinical conversational agent};
\pic at (6.90,0.66) {patient};
\node[leg] at (7.15,0.62) {Patient};

\node[head] (hC) at (1.05,0.10) {Call C};
\node[sub, anchor=west] at ($(hC.east)+(0.16,0)$) {the second ``No.'' escalates};
\draw[labelgrey!30,line width=0.4pt] (0,-0.12) -- (4.3,-0.12);

\node[agentbub] (c1) at (0.42,-0.40)
  {Now I'd like to ask about any discomfort you might be feeling. Are you experiencing any pain in your eye at the moment?};
\pic at ($(c1.north west)+(-0.26,-0.20)$) {agent};

\node[coolbub] (c2) at ($(c1.south west)+(3.88,-0.16)$) {\bfseries ``No.''};
\pic at ($(c2.north east)+(0.26,-0.18)$) {patient};
\node[pill,fill=coolline] at ($(c2.west)+(-0.30,0)$) {not frustrated};

\node[sub] (cgap) at ($(c2.south west)+(-1.55,-0.22)$) {\itshape eighteen turns later};

\node[agentbub] (c3) at ($(cgap.south west)+(-1.30,-0.14)$)
  {Let's summarize what you've told me so far. You had cataract surgery on one eye, and your eye isn't currently red. You're not experiencing any pain\ldots};
\pic at ($(c3.north west)+(-0.26,-0.20)$) {agent};

\node[hotbub] (c4) at ($(c3.south west)+(3.88,-0.16)$) {\bfseries ``No.''};
\pic at ($(c4.north east)+(0.26,-0.18)$) {patient};
\node[pill,fill=hotline] at ($(c4.west)+(-0.42,0)$) {frustrated};

\node[head] (hD) at (7.85,0.10) {Call D};
\node[sub, anchor=west] at ($(hD.east)+(0.16,0)$) {frustration the wording does not carry};
\draw[labelgrey!30,line width=0.4pt] (6.8,-0.12) -- (11.1,-0.12);

\node[agentbub] (d1) at (7.22,-0.40)
  {So you're not seeing any flashing lights now. Do you get pain when you're looking at bright lights?};
\pic at ($(d1.north west)+(-0.26,-0.20)$) {agent};

\node[coolbub] (d2) at ($(d1.south west)+(3.88,-0.16)$) {\bfseries ``Nope.''};
\pic at ($(d2.north east)+(0.26,-0.18)$) {patient};
\node[pill,fill=coolline] at ($(d2.west)+(-0.30,0)$) {not frustrated};

\node[sub] (dgap) at ($(d2.south west)+(-1.55,-0.22)$) {\itshape nineteen turns later};

\node[agentbub] (d3) at ($(dgap.south west)+(-1.30,-0.14)$)
  {\ldots eight weeks can feel like a long time. Do you think you'd like to speak to a doctor or nurse about how you're feeling\ldots};
\pic at ($(d3.north west)+(-0.26,-0.20)$) {agent};

\node[hotbub, text width=32mm] (d4) at ($(d3.south west)+(3.88,-0.16)$)
  {\bfseries ``Eight weeks doesn't feel it is a long time.''};
\pic at ($(d4.north east)+(0.26,-0.18)$) {patient};
\node[pill,fill=hotline] at ($(d4.west)+(-0.42,0)$) {frustrated};

\node[verdict] at ($(d4.south east)+(0,-0.16)$)
  {given the audio, both frontier models call this correctly;\\
   given the transcript, both call it not frustrated};

\end{tikzpicture}
  \caption{Two ways frustration escapes the transcript, both mid-call and both from real calls. \emph{Call C}: the patient answers ``No.'' to a question about pain and ``No.'' again eighteen turns later, and the two carry opposite labels, so the same single word is both. \emph{Call D}: a longer answer whose wording reads as a mild dismissal is labelled frustrated, and both frontier models get it right from the audio while both call it not frustrated from the transcript. Agent prompts are elided where long; patient turns are complete and verbatim.}
  \label{fig:escalation}
\end{figure}

The same difficulty appears mid-call, and in two different shapes.
Figure~\ref{fig:escalation} shows both, from real calls.

Call C is the shortest possible answer. The patient says ``No.'' to a question
about pain, which annotators label not frustrated, and ``No.'' again eighteen
turns later as the agent begins reciting a summary, which they label
frustrated. The two turns are the same single word, so nothing in the
transcript separates them.

Call D is a longer answer whose wording carries no sign of frustration. Asked
whether they would like a clinician to call about a six-to-eight week wait, the
patient replies ``Eight weeks doesn't feel it is a long time.'' Read as text
that is a mild dismissal of the concern; annotators hearing it label it
frustrated. Both frontier models label it correctly from the audio and both
label it not frustrated from the transcript, so the frustration is present in
the delivery and absent from the words.

Both examples sit well before the sign-off, where an escalation could still
change the call.

\subsection{Detecting the onset}
\label{app:onset}

The most consequential turn in
a call is the one where frustration first becomes audible, since that is when
intervention is still cheap. We isolated the onset of each call---the first turn
annotators labelled frustrated---and asked whether each modality cleared it.
\textbf{Text detected only two of the six onsets and confidently cleared three (Gemini 2.5 Flash) or four (Gemini 2.5 Pro);
audio detected four (Flash) or three (Pro) and confidently cleared two.} The two the text
detected are the two that state frustration in words, ``You misunderstand.''
and ``are you going to repeat this all the time?''; the two audio cleared are
``No, it was on the left eye.'' and ``no''. Six calls is an illustration rather than an estimate, and we report it as
one, but the direction is unambiguous and it is the operational shape of the
result: a transcript-based monitor discards the first sign of frustration and
reports the call as calm.

The six onsets are worth reading individually rather than as a rate, both because
six is too few to estimate one and because the turns differ in the way that
matters here.

\begin{figure}[htbp]
  \centering
  \includegraphics[width=0.82\linewidth]{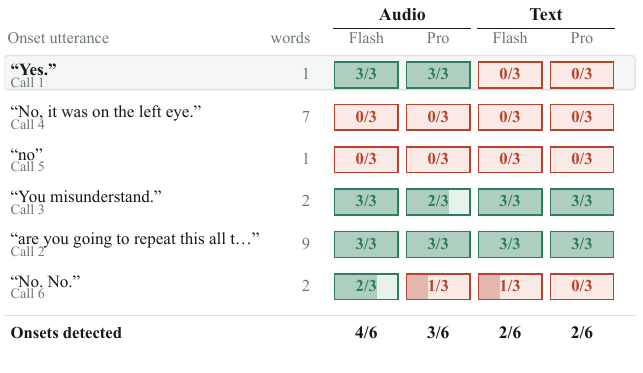}
  \caption{The first turn in each call that annotators labelled frustrated, and
    whether each condition detected it. Four of the six onsets are five words or
    fewer.}
  \label{fig:onset}
\end{figure}

Detection can also be read along the call rather than at a single turn, which
shows how far past the onset a condition first responds.

\begin{figure}[htbp]
  \centering
  \includegraphics[width=0.82\linewidth]{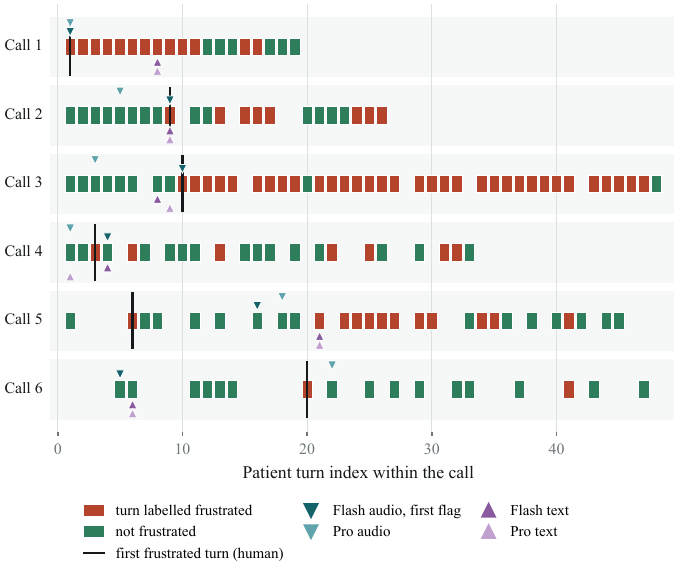}
  \caption{Detection timeline. Each row is one call, each block a patient turn
    coloured by its human label; the rule marks the first frustrated turn and the
    markers show where each condition first returns \emph{frustrated}---audio
    above the row, text below. In the top call both audio conditions fire on the
    onset turn while both text conditions fire seven turns later. Gaps between
    blocks are turns outside the balanced sample, so a first-flag position is
    first among sampled turns rather than first in the call.}
  \label{fig:timeline}
\end{figure}

\section{The 30 calls, measure by measure}
\label{app:thirty}

\begin{figure}[htbp]
  \centering
  \includegraphics[width=\linewidth]{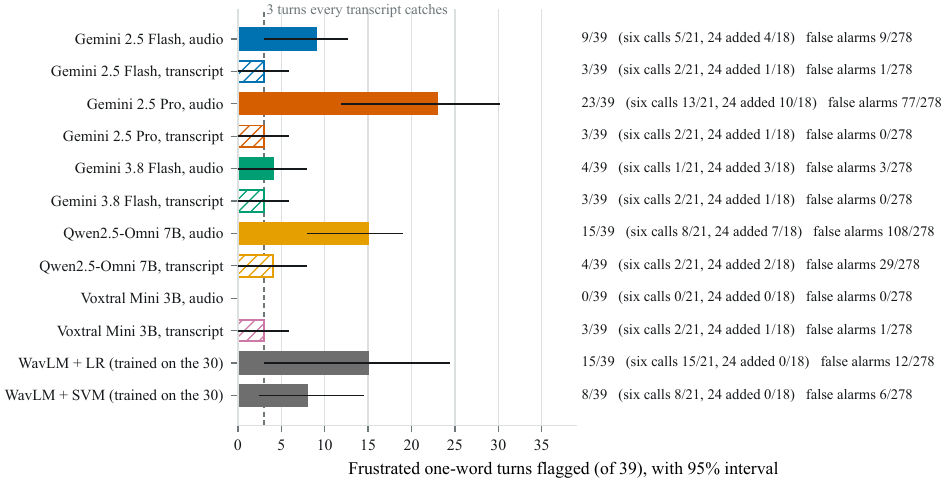}
  \caption{Frustrated one-word turns flagged, out of \num{39} in the 30 calls, for each model and input and for the two WavLM probes trained on the 30 calls, with \SI{95}{\percent} bootstrap intervals over calls. Beside each bar: the split between the six and the 24 calls, and false alarms among the \num{278} one-word turns that were not frustrated. The dashed line marks the three turns every transcript flags.}
  \label{fig:thirtyoneword}
\end{figure}

\begin{figure}[htbp]
  \centering
  \includegraphics[width=\linewidth]{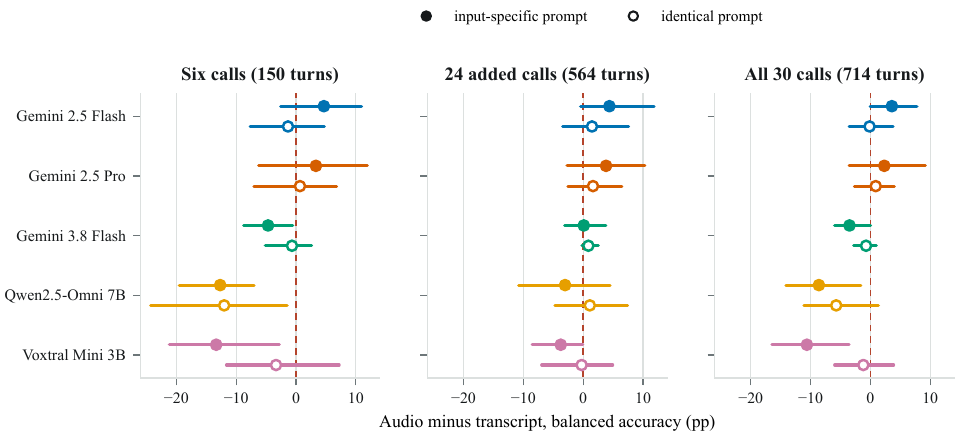}
  \caption{Audio minus transcript balanced accuracy for the five models on the six, the 24 and all 30 calls, under the input-specific prompts (filled) and the identical prompt (hollow), with \SI{95}{\percent} bootstrap intervals over calls.}
  \label{fig:promptgap}
\end{figure}

Table~\ref{tab:thirty} gives each measure of Sections~\ref{sec:c1}--\ref{sec:c3} on the six calls and on the 30 calls, for the three original models and for the two added ones. The corpus comparison, telephony, context, the agent's preceding turn and onset are not repeated on the 30 calls, because they need the acted set or experiments run only on the six. Calls vary more than conditions do: within a condition, balanced accuracy has a standard deviation across calls of \SIrange{7}{19}{pp}, against \SI{5.1}{pp} across the ten model-and-input conditions. Figure~\ref{fig:thirtylengthbias} shows predicted-frustration rate by length on the 24 added calls, whose turns keep their natural prevalence. Figure~\ref{fig:thirtyoneword} gives the frustrated one-word turns each model and input flags, and Figure~\ref{fig:promptgap} the audio-minus-transcript difference for each model under each prompt (Section~\ref{sec:instruction}).

\begin{table}[htbp]
  \caption{Measures of Sections~\ref{sec:c1}--\ref{sec:c3} on the six calls and the 30 calls. Balanced accuracy in percent unless stated; intervals are \SI{95}{\percent} bootstrap intervals over calls, so a few six-call values differ slightly from the main text. Clearing by length counts one-word turns, where Section~\ref{sec:c2} counts turns of one or two words. The prediction spread is shown on the 24 added calls alone.}
  \label{tab:thirty}
  \centering
  \footnotesize
  \setlength{\tabcolsep}{3pt}
  \begin{tabular}{@{}p{3.4cm}p{3.0cm}p{3.6cm}p{3.0cm}@{}}
    \toprule
    Measure & Six calls, 3 models & 30 calls, same 3 models & 3.8 Flash $\cdot$ Voxtral, 30 calls \\
    \midrule
    Share of turns of 1--2 words & \SI{36.7}{\percent} of the balanced 150 & \SI{53.4}{\percent}; \SI{57.8}{\percent} on the 24 & --- \\
    \addlinespace[2pt]
    One-word transcript balanced accuracy & 47.1--54.8 & 49.9--53.8; lower bounds 44.6--50.0 & 53.8 $\cdot$ 53.7 \\
    \addlinespace[2pt]
    Frustrated one-word turns flagged by the transcript & 2 of 21 for each model, the same 2 & 3, 3 and 4 of 39 & 3 of 39 each; all five share the same 3 \\
    \addlinespace[2pt]
    One-word audio balanced accuracy & Flash 60.0, Pro 61.7, Qwen 42.1 & Flash 59.9 [53.3, 63.9], Pro 65.6 [51.9, 74.7], Qwen 49.8 & 54.6 [49.7, 59.8] $\cdot$ 50.0 \\
    \addlinespace[2pt]
    Word count only & 58.0 [50.5, 67.5] & 57.3 [52.8, 62.0]; Qwen audio 55.1 & Voxtral audio 54.4 \\
    \addlinespace[2pt]
    Spread of the transcript's predicted rate across lengths & 50.9--88.1 pp against true 39.4 & On the 24: 26.3--30.9 pp against true 21.2 & 17.3 $\cdot$ 40.4 \\
    \addlinespace[2pt]
    Frustrated share of confidently cleared turns, text against audio & Frontier pair: \SI{37.9}{\percent} of 87 against \SI{17.8}{\percent} of 45; $+20.2$ [6.3, 30.7] & \SI{13.0}{\percent} of 578 against \SI{6.9}{\percent} of 333; $+6.1$ [2.2, 10.4]; 24 alone $+3.3$ [0.3, 6.5] & \SI{13.6}{\percent} against \SI{14.5}{\percent} $\cdot$ \SI{13.6}{\percent} against \SI{17.7}{\percent} \\
    \addlinespace[2pt]
    Wrongly cleared turns that are one word & 18 of 33; \SI{28}{\percent} of frustrated turns are one word & 35 of 75; \SI{28}{\percent} & --- \\
    \addlinespace[2pt]
    Transcript misses / false alarms on short turns & Gemini 23 / 0--1; Qwen 21 / 5 & Gemini 47 / 2; Qwen 44 / 38 & 47 / 0 $\cdot$ 48 / 1 \\
    \addlinespace[2pt]
    Audio minus text & Frontier pair $+4.0$ [$-2.4$, 9.3]; Qwen $-12.7$ [$-19.5$, $-7.0$] & Frontier pair $+3.0$ [$-1.0$, 7.5] (Flash $+3.6$, Pro $+2.3$); Qwen $-8.6$ [$-14.1$, $-1.6$] & $-3.5$ [$-6.0$, 0.0] $\cdot$ $-10.6$ [$-16.5$, $-3.6$] \\
    \addlinespace[2pt]
    Oracle / vote across models and inputs & 78.0 / 66.0 & 89.8 / 67.1; Flash audio 70.2 & --- \\
    \bottomrule
  \end{tabular}
\end{table}

\section{Qualitative examples}
\label{app:cases}

Every rationale below is quoted verbatim from the run that produced the verdict
printed beside it. Turn length is given in words; the ground truth is the majority
of three annotators.

\subsection{Silent clearing}

Both models, reading the transcript, clear a frustrated turn without hesitation. Nothing downstream would flag these. \num{33} such turns occur in the evaluation set.

\begin{casebox}
\casehead{``No.''}{frustrated}{1}
\begin{tabular}{@{}p{0.155\linewidth}p{0.135\linewidth}p{0.63\linewidth}@{}}
Flash / voice & \cross\,calm & \emph{The speaker's single word 'Yeah' is delivered with a neutral to slightly agreeable tone. There is no discernible tension, abruptness, or downward inflection in pitch that would suggest frustration. The pacing is quick, but without other vocal cues, it doesn't indicate an emotional state of frustration.} \\[3pt]
Flash / words & \cross\,calm & \emph{The patient's response 'No' is a single, short, and direct negation. It does not contain any emotionally charged language, exclamations, or elaborations that would suggest frustration. While it could be delivered with a frustrated tone, based on the words alone, it is a neutral answer.} \\[3pt]
Pro / voice & \cross\,calm & \emph{The patient speaks a single word in a neutral, calm tone. The pitch is steady and in a normal range. There are no vocal cues indicating frustration, such as a sigh, a sharp tone, or vocal strain.} \\[3pt]
Pro / words & \cross\,calm & \emph{The user's response is a single word, 'No.' Based on the text alone, this is a simple, direct answer to a question. There are no additional words, phrases, or exclamations to indicate frustration or any other emotional state.} \\[3pt]
\end{tabular}
\end{casebox}

\begin{casebox}
\casehead{``no''}{frustrated}{1}
\begin{tabular}{@{}p{0.155\linewidth}p{0.135\linewidth}p{0.63\linewidth}@{}}
Flash / voice & \cross\,calm & \emph{The patient's tone is relatively neutral and calm when saying 'hello'. There is a slight upward inflection at the end, which is common in a polite greeting, but no discernible harshness, abruptness, or vocal tension that would suggest frustration. The pacing is normal for a single word utterance.} \\[3pt]
Flash / words & \cross\,calm & \emph{The word "no" is a neutral, monosyllabic response that does not inherently convey frustration. There are no additional words, phrasing, or content in this turn to suggest anger, annoyance, or other markers of frustration.} \\[3pt]
Pro / voice & \cross\,calm & \emph{The patient's tone is neutral and even-keeled. The word 'hello' is delivered with a slight upward inflection, which is standard for answering a phone and does not indicate any stress, agitation, or impatience. The pacing is normal and the pitch is within a calm range.} \\[3pt]
Pro / words & \cross\,calm & \emph{The single word 'no' is a direct answer and does not, by itself, contain any linguistic markers of frustration. It is a neutral response.} \\[3pt]
\end{tabular}
\end{casebox}

\subsection{Delivery recovers the turn}

The words are identical to turns that are genuinely calm; the delivery is not, and both models act on it. \num{6} such turns occur in the evaluation set.

\begin{casebox}
\casehead{``No.''}{frustrated}{1}
\begin{tabular}{@{}p{0.155\linewidth}p{0.135\linewidth}p{0.63\linewidth}@{}}
Flash / voice & \tick\,frustrated & \emph{The patient delivers the single word 'No' with an abrupt, firm, and somewhat sharp tone, without any hesitation. This vocal quality suggests a lack of patience or irritation rather than a neutral or calm response.} \\[3pt]
Flash / words & \cross\,calm & \emph{The word 'No' is a simple, direct negation. There is no accompanying phrasing, exclamation, or strong language that would indicate frustration. By itself, it is emotionally neutral and simply answers a perceived question or statement.} \\[3pt]
Pro / voice & \tick\,frustrated & \emph{The single word 'No' is delivered with a sharp, clipped, and abrupt tone. The pacing is very fast for a single syllable, and the pitch is low and forceful, indicating impatience and annoyance rather than a neutral or thoughtful response.} \\[3pt]
Pro / words & \cross\,calm & \emph{The single word 'No' is a direct, neutral, and unembellished response. Standing alone without any further context or emotional language, it does not provide any evidence of frustration. It is simply a negative answer.} \\[3pt]
\end{tabular}
\end{casebox}

\begin{casebox}
\casehead{``No, no.''}{frustrated}{2}
\begin{tabular}{@{}p{0.155\linewidth}p{0.135\linewidth}p{0.63\linewidth}@{}}
Flash / voice & \tick\,frustrated & \emph{The patient's tone is somewhat sharp and abrupt, delivering the two 'no's in quick succession. This quick, emphatic delivery, combined with a slightly dismissive quality in the voice, suggests a feeling of impatience or annoyance, indicating frustration.} \\[3pt]
Flash / words & \cross\,calm & \emph{The phrase "No, no" is a simple and emphatic denial or negation. It does not contain any exclamatory language, strong negative adjectives, or explicit expressions of annoyance, exasperation, or distress that would indicate frustration. The repetition of "no" serves to add emphasis to the denial, which is a neutral linguistic function.} \\[3pt]
Pro / voice & \tick\,frustrated & \emph{The patient's tone is sharp and clipped. The two words are spoken quickly and abruptly, which conveys a sense of impatience or annoyance. The delivery is forceful, not calm or neutral, which points toward frustration.} \\[3pt]
Pro / words & \cross\,calm & \emph{The user is simply saying "No, no." which is a direct negation. The words themselves do not contain any language that indicates frustration, impatience, or annoyance. Without any other context or explicit negative wording, the statement is neutral.} \\[3pt]
\end{tabular}
\end{casebox}

\subsection{Delivery misleads}

The reverse case. Hearing the turn moves both models to the wrong answer. \num{3} such turns occur in the evaluation set.

\begin{casebox}
\casehead{``No!''}{frustrated}{1}
\begin{tabular}{@{}p{0.155\linewidth}p{0.135\linewidth}p{0.63\linewidth}@{}}
Flash / voice & \cross\,calm & \emph{The patient says 'No' with a firm, clear, and steady tone. The pitch is relatively neutral, and there is no discernible strain, raised volume, or erratic pacing that would suggest frustration. It sounds like a definitive, straightforward negative response rather than an expression of annoyance or anger.} \\[3pt]
Flash / words & \tick\,frustrated & \emph{The patient uses a single, absolute negation 'No!' followed by an exclamation mark. This combination of a direct refusal and an emphatic punctuation suggests a strong emotional response, likely indicating frustration, exasperation, or strong disagreement rather than a neutral denial.} \\[3pt]
Pro / voice & \cross\,calm & \emph{The patient says a single word, 'No,' in a low-pitched, even-toned, and calm manner. The delivery is not sharp, clipped, or high-pitched, which would typically be associated with frustration. The tone is neutral and declarative.} \\[3pt]
Pro / words & \tick\,frustrated & \emph{The response is a single, abrupt, and emphatic word, "No!". The use of an exclamation mark indicates a strong emotional charge rather than a simple, neutral answer. This type of forceful, monosyllabic negation is a common way to express frustration or exasperation.} \\[3pt]
\end{tabular}
\end{casebox}

\subsection{Neither recovers the turn}

The common one-word case: every condition clears a frustrated turn. \num{7} such turns occur in the evaluation set.

\begin{casebox}
\casehead{``No.''}{frustrated}{1}
\begin{tabular}{@{}p{0.155\linewidth}p{0.135\linewidth}p{0.63\linewidth}@{}}
Flash / voice & \cross\,calm & \emph{The speaker's single word 'Yeah' is delivered with a neutral to slightly agreeable tone. There is no discernible tension, abruptness, or downward inflection in pitch that would suggest frustration. The pacing is quick, but without other vocal cues, it doesn't indicate an emotional state of frustration.} \\[3pt]
Flash / words & \cross\,calm & \emph{The patient's response 'No' is a single, short, and direct negation. It does not contain any emotionally charged language, exclamations, or elaborations that would suggest frustration. While it could be delivered with a frustrated tone, based on the words alone, it is a neutral answer.} \\[3pt]
Pro / voice & \cross\,calm & \emph{The patient speaks a single word in a neutral, calm tone. The pitch is steady and in a normal range. There are no vocal cues indicating frustration, such as a sigh, a sharp tone, or vocal strain.} \\[3pt]
Pro / words & \cross\,calm & \emph{The user's response is a single word, 'No.' Based on the text alone, this is a simple, direct answer to a question. There are no additional words, phrases, or exclamations to indicate frustration or any other emotional state.} \\[3pt]
\end{tabular}
\end{casebox}

\subsection{The word itself is not recovered}

On a handful of one-word turns the model does not establish which word was
spoken, and reasons confidently about a word that was never said. The transcript
of the clip, produced independently, is given for comparison.

\begin{casebox}
\casehead{``No.''}{frustrated}{1}
\casenote{Independent transcription of this clip: ``No.''.}
\begin{tabular}{@{}p{0.155\linewidth}p{0.135\linewidth}p{0.63\linewidth}@{}}
Flash / voice & \cross\,calm & \emph{The speaker's single word 'Yeah' is delivered with a neutral to slightly agreeable tone. There is no discernible tension, abruptness, or downward inflection in pitch that would suggest frustration. The pacing is quick, but without other vocal cues, it doesn't indicate an emotional state of frustration.} \\[3pt]
Flash / words & \cross\,calm & \emph{The patient's response 'No' is a single, short, and direct negation. It does not contain any emotionally charged language, exclamations, or elaborations that would suggest frustration. While it could be delivered with a frustrated tone, based on the words alone, it is a neutral answer.} \\[3pt]
Pro / voice & \cross\,calm & \emph{The patient speaks a single word in a neutral, calm tone. The pitch is steady and in a normal range. There are no vocal cues indicating frustration, such as a sigh, a sharp tone, or vocal strain.} \\[3pt]
Pro / words & \cross\,calm & \emph{The user's response is a single word, 'No.' Based on the text alone, this is a simple, direct answer to a question. There are no additional words, phrases, or exclamations to indicate frustration or any other emotional state.} \\[3pt]
\end{tabular}
\end{casebox}

\begin{casebox}
\casehead{``No.''}{frustrated}{1}
\casenote{Independent transcription of this clip: ``No.''.}
\begin{tabular}{@{}p{0.155\linewidth}p{0.135\linewidth}p{0.63\linewidth}@{}}
Flash / voice & \cross\,calm & \emph{The patient's single word 'Hello' is delivered with a neutral and even tone, without any raised pitch, abruptness, or vocal tension that would indicate frustration.} \\[3pt]
Flash / words & \cross\,calm & \emph{The patient's response 'No.' is a simple, direct negation. Without any additional context, inflection (which we cannot infer from text alone), or other words, there is nothing in this single word to indicate frustration. It serves as a neutral answer.} \\[3pt]
Pro / voice & \tick\,frustrated & \emph{The patient says a single word, "No," with a sharp, clipped, and slightly forceful tone. The abruptness of the utterance and the slightly downward, declarative pitch suggest impatience and annoyance rather than a calm or neutral state.} \\[3pt]
Pro / words & \cross\,calm & \emph{The single word "No" is a direct and simple answer. Without any additional context or emotional language, it does not inherently convey frustration. It is a neutral response to a presumed yes/no question.} \\[3pt]
\end{tabular}
\end{casebox}

\end{document}